%% file: main.tex
\documentclass{article}

\usepackage{arxiv}

\usepackage[utf8]{inputenc}
\usepackage[T1]{fontenc}
\usepackage[scaled=.98]{XCharter}
\usepackage[scaled=.95]{helvet}
\usepackage[scaled=1.1]{zlmtt}
\usepackage{amsmath}
\usepackage{amssymb}
\usepackage[uprightscript,charter,vvarbb,scaled=1.05]{newtxmath}
\usepackage[hyphens]{url}
\usepackage[table]{xcolor}
\definecolor{LinkBlue}{HTML}{2457C5}
\definecolor{AbstractGray}{HTML}{F3F5F7}
\definecolor{AbstractBorder}{HTML}{DDE3EA}
\usepackage[
  colorlinks=true,
  linkcolor=LinkBlue,
  citecolor=LinkBlue,
  urlcolor=LinkBlue
]{hyperref}
\usepackage{graphicx}
\usepackage{fontawesome5}
\usepackage{booktabs}
\usepackage{tabularx}
\usepackage{microtype}
\usepackage{enumitem}
\usepackage{etoolbox}
\usepackage[numbers,sort&compress]{natbib}
\usepackage{doi}
\usepackage[most]{tcolorbox}

\newtcolorbox{preprintabstract}{
  enhanced,
  breakable,
  colback=AbstractGray,
  colframe=AbstractBorder,
  boxrule=0.35pt,
  arc=3pt,
  left=12pt,
  right=12pt,
  top=5pt,
  bottom=9pt,
  before skip=10pt,
  after skip=16pt,
  title={Abstract},
  fonttitle=\large\bfseries\sffamily,
  coltitle=black,
  colbacktitle=AbstractGray,
  halign title=center,
  titlerule=0pt
}

\renewenvironment{abstract}
  {\begin{preprintabstract}\fontsize{10pt}{12.4pt}\selectfont}
  {\end{preprintabstract}}

\makeatletter
\renewenvironment{table}
  {\@float{table}}
  {\end@float}
\makeatother

\pdftrailerid{}

\input{generated/evidence-macros.tex}

\usepackage{array}
\usepackage{caption}
\usepackage{placeins}
\makeatletter
\renewenvironment{table*}{\@float{table}}{\end@float}
\renewenvironment{figure*}{\@float{figure}}{\end@float}
\makeatother

\def\eqref#1{equation~\ref{#1}}

\newcommand{\paperstatus}{Preprint}
\newcommand{\dezhicorrespondenceemail}{dezhiran@pku.edu.cn}
\newcommand{\taocorrespondenceemail}{taoxie@pku.edu.cn}
\title{\normalfont\sffamily\bfseries Backward-State Policy Is Part of the\\Learning Algorithm}
\newcommand{\authornamefont}{\fontsize{11.8pt}{14.2pt}\selectfont\sffamily\bfseries}
\newcommand{\affiliationfont}{\fontsize{10.2pt}{12.4pt}\selectfont\normalfont}
\newlength{\authorrowbreakheight}
\makeatletter
\patchcmd{\@maketitle}
  {\begin{tabular}[t]{c}\bf\rule{\z@}{24\p@}\ignorespaces}
  {\begin{tabular}[t]{c}\authornamefont\rule{\z@}{24\p@}\ignorespaces}
  {}{\PackageWarning{arxiv-paper-template}{Could not patch the first author-row break}}
\patchcmd{\@maketitle}
  {\begin{tabular}[t]{c}\bf\rule{\z@}{24\p@}\ignorespaces}
  {\begin{tabular}[t]{c}\authornamefont\rule{\z@}{\authorrowbreakheight}\ignorespaces}
  {}{\PackageWarning{arxiv-paper-template}{Could not patch the second author-row break}}
\makeatother
\newcommand{\resourcelink}[2]{%
  \href{#1}{\normalfont\mdseries\nolinkurl{#2}}%
}

\author{
  \authornamefont
  \textbf{Shuxiao Xie}\textsuperscript{1,2}\thanks{Equal contribution.}
  \quad
  \textbf{Shuyang Xie}\textsuperscript{3}\footnotemark[1]
  \AND
  \textbf{Dezhi Ran}\textsuperscript{1}\thanks{Corresponding authors:
  \resourcelink{mailto:\dezhicorrespondenceemail}{\dezhicorrespondenceemail} and
  \resourcelink{mailto:\taocorrespondenceemail}{\taocorrespondenceemail}.}
  \quad
  \textbf{Wei Yang}\textsuperscript{2}
  \quad
  \textbf{Tao Xie}\textsuperscript{1,2,4,5}\footnotemark[2]
  \\[0.7em]
  \affiliationfont
  \textsuperscript{1}Beijing Tongming Lake Information Technology Application
  Innovation Center (TLAIC), China
  \\
  \textsuperscript{2}Fudan University Institute of Systems for Advanced Computing, China
  \\
  \textsuperscript{3}Harbin Institute of Technology, China
  \\
  \textsuperscript{4}Key Lab of HCST (PKU), MOE; SCS, Peking University, Beijing, China
  \\
  \textsuperscript{5}Shanghai Institute of Systems for Open Computing, China
}

\date{}
\renewcommand{\headeright}{\paperstatus}
\renewcommand{\undertitle}{\paperstatus}
\renewcommand{\shorttitle}{Backward-State Policy}
\hypersetup{pdftitle={Backward-State Policy Is Part of the Learning Algorithm},pdfsubject={Public preprint},pdfauthor={Shuxiao Xie, Shuyang Xie, Dezhi Ran, Wei Yang, Tao Xie}}

\begin{document}
\maketitle

\begin{abstract}
Low-precision training rounds tensors that the backward pass reads again, often for several gradients; each use can read the forward's rounded value, the original, or a new random rounding. This backward-state policy looks like a memory and precision detail, settled by copy accuracy and final loss. We argue that it is part of the learning algorithm, and that neither check shows whether it is right. Copy accuracy does not decide the outcome: in three pairs of 390M runs with an emulated FP8 backward, training fails when attention's backward reuses the forward's rounded output and succeeds with a new rounding from the same distribution. Even the most accurate copy, the original itself, can be wrong by our reference: the gradient of the forward pass as it actually ran, with gradients passed through rounding unchanged. For example, a normalization output stored in low precision feeds two gradients: the gain's gradient needs the original, but the next layer's weight gradient needs the rounded value that layer multiplied. Final loss, the other check, does not rule out the error of reading the original for both: it persists in models trained with such a store, while planned loss comparisons stay within a margin fixed in advance. We therefore derive from this reference which value each use must read, or which substitute gives the same gradient on average with the forward held fixed, and check these per-use requirements on single operators, without training. In three tests using PyTorch and Transformer Engine, the requirements predicted beforehand whether reuse changes what the backward computes on average relative to an independent copy, and every prediction held. Backward-state policy is thus part of the learning algorithm: it should be specified and
checked use by use, not settled by copy accuracy and final loss.
\end{abstract}

\IfFileExists{sections/01-introduction.tex}{\input{sections/01-introduction}}{}
\IfFileExists{sections/02-outcome.tex}{\input{sections/02-outcome}}{}
\IfFileExists{sections/03-uses.tex}{\input{sections/03-uses}}{}
\IfFileExists{sections/04-contract.tex}{\input{sections/04-contract}}{}
\IfFileExists{sections/05-conclusion.tex}{\input{sections/05-conclusion}}{}

\bibliographystyle{plainnat}
\bibliography{references}

\appendix
\IfFileExists{sections/A-related.tex}{\input{sections/A-related}}{}
\IfFileExists{sections/B-theory.tex}{\input{sections/B-theory}}{}
\IfFileExists{sections/C-contract.tex}{\input{sections/C-contract}}{}
\IfFileExists{sections/D-setup.tex}{\input{sections/D-setup}}{}
\IfFileExists{sections/E-evidence.tex}{\input{sections/E-evidence}}{}

\end{document}

%% file: generated/evidence-macros.tex
\newcommand{\NSOneSevenZeroEConsETwo}{0.0018}
\newcommand{\NSOneSevenZeroEConsEThree}{0.0001}
\newcommand{\NSOneSevenZeroEFpEightDpaExcursionStep}{4900}
\newcommand{\NSOneSevenZeroEFpEightDpaLossAtFourNineZeroZero}{7.40}
\newcommand{\NSOneSevenZeroEFpEightDpaLossAtFourNineNineNine}{3.47}
\newcommand{\NSOneSevenZeroEFpEightDpaRunningMinBeforeFourNineZeroZero}{2.96}
\newcommand{\NSOneEightOneIIEightLM}{208.0}
\newcommand{\NSOneEightOneIIEightETwo}{+0.0017}
\newcommand{\NSOneEightOneIIEightEThree}{+0.0008}
\newcommand{\NSOneEightOneIIEightBLM}{290.0}
\newcommand{\NSOneEightOneIIEightBETwo}{+0.0024}
\newcommand{\NSOneEightOneIIEightBEThree}{+0.0023}
\newcommand{\NSOneEightOneIEightRmsRatio}{1.39}
\newcommand{\NSOneEightOneIEightCorrIEightOEight}{0.0005}
\newcommand{\NSOneEightFourVPassLM}{26368.0}
\newcommand{\NSOneEightFourVPassETwo}{+0.0876}
\newcommand{\NSOneEightFourVPassEThree}{+0.0454}
\newcommand{\NSOneEightFourVSrLM}{166.0}
\newcommand{\NSOneEightFourVSrETwo}{+0.0009}
\newcommand{\NSOneEightFourVSrEThree}{-0.0010}
\newcommand{\NSOneEightFourVSrTwoLM}{218.0}
\newcommand{\NSOneEightFourVSrTwoETwo}{+0.0010}
\newcommand{\NSOneEightFourVSrTwoEThree}{-0.0010}
\newcommand{\NSOneEightFourVRtnLM}{11520.0}
\newcommand{\NSOneEightFourVRtnETwo}{+0.0282}
\newcommand{\NSOneEightFourVRtnEThree}{+0.0217}
\newcommand{\NSOneEightFourVSrRmsRatioLTen}{1.35}
\newcommand{\NSOneEightFourVSrCorrLTen}{0.0009}
\newcommand{\NSOneEightFourVRtnRmsRatioLTen}{0.96}
\newcommand{\NSOneEightFourVRtnCorrLTen}{0.7536}
\newcommand{\NSOneEightFourVBrefLM}{197.0}
\newcommand{\NSOneNineOneBaseZ}{30.59}
\newcommand{\NSOneNineOneOutFZ}{0.06}
\newcommand{\NSOneNineOneInFZ}{33.30}
\newcommand{\NSOneNineOneOutFPctOfBase}{0.19}
\newcommand{\NSOneNineOneInFPctOfBase}{106.12}
\newcommand{\NSOneNineOneOutFMaskAgreement}{0.4998}
\newcommand{\NSOneNineTwoLSOneZ}{0.68}
\newcommand{\NSOneNineTwoLSOnePctOfB}{0.17}
\newcommand{\NSOneNineTwoLSTwoRelErr}{0.004385}
\newcommand{\NSOneNineTwoLSThreeRelErr}{-0.002878}
\newcommand{\NSOneNineTwoLSFourZ}{2.033}
\newcommand{\NSOneNineTwoLSFourPctOfA}{0.91}
\newcommand{\NSOneNineTwoLSSixRelErr}{6.08e-14}
\newcommand{\NSOneNineTwoLSSixZ}{29.94}
\newcommand{\NSOneNineThreeCCellAMaxDbdV}{27/640}
\newcommand{\NSOneNineThreeCCellBMaxDbdV}{17/320}
\newcommand{\NSOneNineThreeCNullMaxDbdV}{0}
\newcommand{\NSOneNineThreeCMaxGapReplay}{17/320}
\newcommand{\NSOneNineThreeCMaxGapRole}{0}
\newcommand{\NSOneEightFourPPoSZeroDEtwo}{0.006281395}
\newcommand{\NSOneEightFourPPoSOneDEtwo}{0.00866544}
\newcommand{\NSOneNineTwoLSFiveCalibRatio}{0.9950207489532267}
\newcommand{\NSOneNineTwoLSFiveACalibZ}{0.29}
\newcommand{\NSOneNineTwoLSFiveACalibPctOfAShared}{0.10}
\newcommand{\NSOneNineThreeASrRefreshOverReplay}{0.4808}
\newcommand{\NSOneNineThreeASymRefreshOverReplay}{0.5598}
\newcommand{\NSOneNineThreeBSrRefreshOverReplay}{0.4222}
\newcommand{\NSOneNineThreeBSymRefreshOverReplay}{0.4254}
\newcommand{\NSTwoOneOneDf}{6}
\newcommand{\NSTwoOneOneDeltaNats}{0.010}
\newcommand{\NSTwoOneOneKSevenDBar}{-0.00425}
\newcommand{\NSOneSevenSixCRZeroSrOneFinalTrain}{7.314}
\newcommand{\NSOneSevenSixCRZeroSrZeroFinalTrain}{3.256}
\newcommand{\NSOneSevenSixCRZeroSrBFinalTrain}{3.256}
\newcommand{\NSTwoOneFiveDoseFiveKSixDBar}{-0.00957}
\newcommand{\NSTwoOneFiveDoseFiveDf}{4}
\newcommand{\NSOneSevenFiveSZeroNativeOnset}{4500}
\newcommand{\NSOneSevenFiveSZeroNativeDVal}{5.097}
\newcommand{\NSOneSevenFiveSOneNativeOnset}{5500}
\newcommand{\NSOneSevenFiveSOneNativeDVal}{5.120}
\newcommand{\NSOneSevenFiveSTwoNativeOnset}{5000}
\newcommand{\NSOneSevenFiveSTwoNativeDVal}{5.225}
\newcommand{\NSOneSevenFiveNNativeFails}{3}
\newcommand{\NSOneEightFourPPoSTwoDEtwo}{0.00817628}
\newcommand{\NSTwoTwoSixLSEps}{0.5}
\newcommand{\NSTwoTwoSixLSTwoRelErr}{0.00067}
\newcommand{\NSTwoTwoSixLSTwoZ}{1418}
\newcommand{\NSTwoTwoSixLSTwoDRelErrLead}{0.0217}
\newcommand{\NSTwoTwoSixLSThreeRelErr}{0.00169}
\newcommand{\NSTwoTwoSixLSThreeZ}{298}
\newcommand{\NSTwoTwoSixLSThreeDRelErrLead}{0.0440}
\newcommand{\NSTwoTwoSixLSOneContrast}{-5.63e-05}
\newcommand{\NSTwoTwoSixLSOneMargin}{0.0130}
\newcommand{\NSTwoTwoSixLSFourBIndep}{-1.20e-03}
\newcommand{\NSTwoTwoSixLSFourMargin}{0.0128}
\newcommand{\NSTwoTwoSixLSFiveACalib}{1.26e-04}
\newcommand{\NSTwoTwoSixLSSixRelErr}{5.0e-14}
\newcommand{\NSTwoTwoSixCKWidth}{512}
\newcommand{\NSTwoTwoSixCKRBaseZ}{65.4}
\newcommand{\NSTwoTwoSixCKIBaseZ}{90.4}
\newcommand{\NSTwoTwoSixCKROutfPctOfBase}{2.90}
\newcommand{\NSTwoTwoSixCKRInfRatio}{0.962}
\newcommand{\NSTwoTwoSixCKIOutfPctOfBase}{0.75}
\newcommand{\NSTwoTwoSixCKIInfRatio}{0.994}
\newcommand{\NSTwoTwoSixCKRMaskAgreeOutf}{0.500}
\newcommand{\NSTwoTwoSixCKRtnBaseZ}{63.6}
\newcommand{\NSTwoTenCSZeroPThreeCount}{16}
\newcommand{\NSTwoTenCSTwoPThreeCount}{16}
\newcommand{\NSTwoSeventeenOnsetSZeroBelowRefSixTwoFourNine}{16}
\newcommand{\NSTwoSeventeenOnsetSOneBelowRefSixTwoFourNine}{16}
\newcommand{\NSOneSevenSixCMNatLabel}{FAILS}
\newcommand{\NSOneSevenSixCMNatLM}{40448}
\newcommand{\NSOneSevenSixCMCnsLM}{198}
\newcommand{\NSOneSevenSixCMMirLabel}{IGNITED}
\newcommand{\NSOneSevenSixCMMirLM}{29696}
\newcommand{\NSOneSevenSixCMMibLabel}{USEFUL}
\newcommand{\NSOneSevenSixCMMibLM}{202}
\newcommand{\NSOneSevenSixCMNatGapETwo}{+2.4273}
\newcommand{\NSOneSevenSixCMNatGapEThree}{+3.7973}
\newcommand{\NSOneSevenSixCMMirGapETwo}{+0.0482}
\newcommand{\NSOneSevenSixCMMirGapEThree}{+0.0366}
\newcommand{\NSOneSevenSixCMMibGapETwo}{+0.0006}
\newcommand{\NSOneSevenSixCMMibGapEThree}{+0.0029}
\newcommand{\NSOneEightSixScaleMeasuredRatio}{256.71}
\newcommand{\NSOneEightSixScalePredictedRmsTwo}{256.50}
\newcommand{\NSOneEightSixScaleAgreement}{1.0009}
\newcommand{\NSOneEightSixCtrlZ}{0.73}
\newcommand{\NSOneEightSixSignResolvedPositive}{0}
\newcommand{\NSOneEightSixSignEResolvedPositive}{24}
\newcommand{\NSTwoTwoFourCIZeroSrZeroGapETwo}{+0.00002}
\newcommand{\NSTwoTwoFourCIZeroSrZeroGapEThree}{+0.0020}
\newcommand{\NSTwoTwoFourCIZeroSrOneSOneOpen}{1787}
\newcommand{\NSTwoTwoFourBOneSZeroEOneOnset}{3600}
\newcommand{\NSTwoTwoFourBOneSTwoEOneOnset}{5400}
\newcommand{\NSTwoTwoSevenGpurowMaxAbsDev}{0}
\newcommand{\NSTwoTwoSevenGpurowEps}{0.00625}
\newcommand{\NSTwoTwoSevenGpurowMeanValue}{0.0625}
\newcommand{\NSTwoTwoSevenGpurowMeanFixed}{0}
\newcommand{\NSTwoTwoSevenGpurowPredMeanValue}{0.0625}
\newcommand{\NSTwoTwoNineRowsPvFourMaxSatShare}{25.2}
\newcommand{\NSTwoTwoNineRowsPbLmNat}{40448}
\newcommand{\NSTwoTwoNineRowsPbLmUnsat}{30208}
\newcommand{\NSTwoTwoNineRowsPbLmSat}{188}
\newcommand{\NSTwoTwoNineRowsPbLmB}{198}
\newcommand{\NSTwoTwoNineRowsPbLmSOneSix}{194}
\newcommand{\NSTwoTwoNineRowsPbETwoNat}{+2.4273}
\newcommand{\NSTwoTwoNineRowsPbEThreeNat}{+3.7973}
\newcommand{\NSTwoTwoNineRowsPbLabelNat}{FAILS}
\newcommand{\NSTwoTwoNineRowsPbETwoUnsat}{+1.2702}
\newcommand{\NSTwoTwoNineRowsPbEThreeUnsat}{+0.3612}
\newcommand{\NSTwoTwoNineRowsPbLabelUnsat}{FAILS}
\newcommand{\NSTwoTwoNineRowsPbETwoSat}{+0.0020}
\newcommand{\NSTwoTwoNineRowsPbEThreeSat}{+0.0019}
\newcommand{\NSTwoTwoNineRowsPbLabelSat}{USEFUL}
\newcommand{\NSTwoTwoNineRowsPbETwoSOneSix}{+0.0015}
\newcommand{\NSTwoTwoNineRowsPbEThreeSOneSix}{-0.0008}
\newcommand{\NSTwoTwoNineRowsPbLabelSOneSix}{USEFUL}
\newcommand{\NSTwoTwoNineImgShiftVal}{-0.0014}
\newcommand{\NSTwoTwoNineImgShiftETwo}{-0.0010}
\newcommand{\NSTwoTwoNineEmulDqMin}{2.4}
\newcommand{\NSTwoTwoNineEmulDqMax}{5.9}
\newcommand{\NSTwoTwoNineEmulDkMin}{2.8}
\newcommand{\NSTwoTwoNineEmulDkMax}{7.6}
\newcommand{\NSOneEightFourPPoSZeroDValDisp}{0.0054}
\newcommand{\NSOneEightFourPPoSOneDValDisp}{0.0076}
\newcommand{\NSOneEightFourPPoSTwoDValDisp}{0.0067}
\newcommand{\NSOneSevenFiveSZeroConsDValDisp}{0.0029}
\newcommand{\NSOneSevenFiveSZeroConsDEtwoDisp}{0.0037}
\newcommand{\NSOneSevenFiveSOneConsDValDisp}{0.0059}
\newcommand{\NSOneSevenFiveSOneConsDEtwoDisp}{0.0071}
\newcommand{\NSOneSevenFiveSTwoConsDValDisp}{0.0052}
\newcommand{\NSOneSevenFiveSTwoConsDEtwoDisp}{0.0061}
\newcommand{\NSTwoThreeZeroOptRRGamma}{1/3}
\newcommand{\NSTwoThreeZeroOptRRExcess}{1/18}
\newcommand{\NSTwoThreeZeroOptRRGGammaAtOpt}{1/2}
\newcommand{\NSTwoThreeZeroOptSSGWAtOpt}{-1/4}
\newcommand{\NSTwoThreeZeroOptOptGamma}{1/2}
\newcommand{\NSTwoThreeZeroOptOptW}{1}
\newcommand{\NSTwoThreeZeroOptJStar}{9/32}
\newcommand{\NSTwoTwoFourBTwoSZeroConsDval}{+0.0075}
\newcommand{\NSTwoTwoFourBTwoSZeroConsDeTwo}{+0.0053}
\newcommand{\NSTwoTwoFourBTwoSTwoConsDval}{+0.0050}
\newcommand{\NSTwoTwoFourBTwoSTwoConsDeTwo}{+0.0046}
\newcommand{\NSTwoThreeFourGpuhTotal}{5856}
\newcommand{\NSTwoThreeFourGpuhNRuns}{112}
\newcommand{\NSTwoThreeFourGpuhNParents}{3}
\newcommand{\NSTwoThreeFourGpuhWorld}{8}
\newcommand{\NSTwoThreeFourGpuhCfgMInitTwentyKMedian}{79}
\newcommand{\NSTwoThreeFourGpuhCfgMInitTwentyKSmallest}{66}
\newcommand{\NSTwoThreeFourGpuhCfgMInitTwentyKLargest}{95}

%% file: sections/01-introduction.tex
%
\section{Introduction}\label{sec:intro}

Large language models are now trained in low precision, with FP8 (8-bit floating point) at scale
\citep{peng2023training,fishman2024scaling,mishra2025recipes} and 4-bit formats following
\citep{castro2025quartet,chmiel2025all}. The forward pass rounds tensors before operations consume them, and the backward
pass reads many of these tensors again, often for several gradients. Keeping them at full precision costs memory, so
implementations decide what the backward receives: FlashAttention saves the attention output
\citep{dao2022flashattention}, activation-compressed training keeps few-bit copies of activations for the backward
\citep{chen2021actnn,liu2022gact}, and checkpointing recomputes parts of the forward \citep{pytorch2026checkpoint}.\footnote{The main text discusses
related work where it bears on each result; Appendix~\ref{sec:app_related} surveys it by area.} Once
a tensor has been rounded, these choices decide which value each gradient reads: the rounded value the forward used, the
original, or a new random rounding of the original. We call each gradient computation that reads a rounded tensor a use,
and the choice made for every use the backward-state policy.

Viewed as ways of keeping a tensor, these choices look like a memory and precision detail, settled by two checks:
whether the copy is close to the original, and whether the final loss looks normal. A use, however, reads its value to
compute one particular gradient, so the value it reads can change the gradient that training follows. We argue that the
backward-state policy is part of the learning algorithm, and that neither check shows whether it is right.

Low-precision attention shows what is at stake. FOG, a study of fully FP8 training, reports that FP8 attention training
diverges on standard architectures and avoids the failure with new architectures \citep{hernndezcano2025fully}. In our
runs, changing only the backward can prevent such a failure. In FOG's 390M setting, the FP8 attention of NVIDIA's
Transformer Engine fails on all \NSOneSevenFiveNNativeFails{} seeds we ran, and replacing only its backward with a BF16
(bfloat16) one removes the failure while keeping the FP8 forward
(Section~\ref{sec:outcome}). In a separate study of BF16 FlashAttention,
\citet{qiu2025why} locate a failure in a correction term of the attention backward, $D=\mathrm{rowsum}(dO\circ O)$,
which reads the saved output $O$ and its incoming gradient $dO$. They trace the term's error to biased rounding in
computing $O$, which their change to the forward's softmax normalization prevents. Attn-QAT, a 4-bit attention method,
instead gives $D$ a separate high-precision attention output \citep{zhang2026attn}. \citeauthor{qiu2025why}'s fix and
Attn-QAT's separate output both make the value that $D$ reads more accurate.

\begin{figure}[t]
\centering
\includegraphics[width=\linewidth]{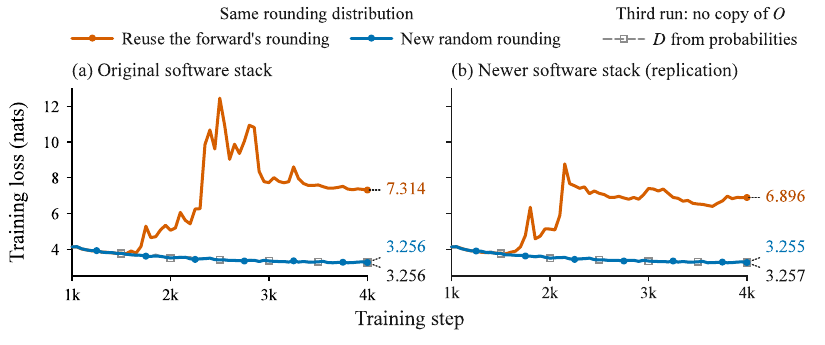}
\caption{\textbf{Same rounding distribution, opposite outcomes.} Training loss of 390M runs with an emulated FP8
attention backward, forked from one checkpoint; all consume a stochastic rounding of the attention output $O$, and within
each panel they differ only in how the correction term $D$ is computed. \textbf{(a)}~The first of three pairs: reusing the forward's rounding
fails (final loss \NSOneSevenSixCRZeroSrOneFinalTrain{}); a new rounding from the same distribution trains
(\NSOneSevenSixCRZeroSrZeroFinalTrain{}), like a control whose $D$ reads no copy of $O$. \textbf{(b)}~The runs of (a),
repeated on newer versions of PyTorch and Transformer Engine.}
\label{fig:pairing}
\end{figure}

If copy accuracy settled the choice, $D$ could read either of two equally accurate copies of $O$ without changing whether
training succeeds. We test this directly, letting the forward round $O$ at random \citep{gupta2015deep} so that a new
rounding from the same distribution is as accurate as the forward's own. The two differ in one respect: only the forward's
rounding went into the loss, so $dO$ can depend on it but not on the new one. In our 390M runs with an emulated FP8
attention backward, training fails when $D$ reuses the forward's rounding and trains when $D$ reads a new rounding from
the same distribution (Figure~\ref{fig:pairing}).

These results show that what the backward reads can decide the outcome. A successful run still leaves open which
gradient it followed and which choice is right. To check whether a policy preserves the gradient of the
computation that produced the loss, we declare a reference: the gradient of the forward pass as it actually ran, with
gradients passed through each rounding unchanged as in the straight-through convention
\citep{bengio2013estimating,yin2019understanding}. Under this reference, each use needs the value its forward operation
used, rounded or original, or a substitute that gives the same gradient on average conditional on the realized forward
and the incoming gradient the use actually receives.

Judged this way, even the most accurate copy can be wrong, because one tensor can feed gradients that need different
values. A normalization output rounded to low precision in the forward feeds two: the gain's gradient needs the
original, because the gain scaled unrounded values, and the next layer's weight gradient needs the rounded value,
because that layer multiplied it. No single value read by both is always right for both. For the probabilities that
FP8 attention rounds before multiplying them by $V$, we prove, under stated assumptions, that no single substitute is
always right on average at both of their uses.

Final loss, the other check, does not reveal such an error either. In our 390M models trained with the rounded output
feeding the next layer and the original read by both gradients, the error persists, yet the final loss stays within a
margin, fixed in advance, of the policy that follows the reference. Loss can show that a policy fails to train; it
cannot show that it keeps the reference.

We therefore check the policy at each use. From the reference we derive per-use requirements, which state the value each
use must read or the substitutes it may read. Each can be checked on its operator without training, using the random
dependencies of the incoming gradient it actually receives. An executable checker tests stated policies against the
requirements on small finite cases. The requirements also predict when reuse matters, the contrast of Figure~\ref{fig:pairing}. Averaged over both
roundings, handing a use the forward's rounding instead of a new one can change what the backward computes only if the
use's incoming gradient depends on that rounding. We registered such predictions before testing three operators of PyTorch and
Transformer Engine, and every one held, including predictions that reuse changes nothing.

%% file: sections/02-outcome.tex
%
\section{What a use reads can decide the outcome}\label{sec:outcome}

\subsection{Same rounding distribution, opposite outcomes}\label{sec:outcome-forks}

Attention's correction term shows why two copies from one rounding distribution need not be interchangeable. The
forward turns scores $S$ into probabilities $P=\mathrm{softmax}(S)$ and outputs $O=PV$. Given the output's gradient
$dO$, the backward forms $dV=P^\top dO$ and $dP=dO\,V^\top$, and the softmax's backward gives $dS=P\circ(dP-D)$ with
$D=\mathrm{rowsum}(P\circ dP)$. FlashAttention computes $D$ from the saved output as $\mathrm{rowsum}(dO\circ O)$, the
same value because $O=PV$, and reads the saved output nowhere else \citep{dao2022flashattention}. When the forward
rounds $O$ before the next layer consumes it, our reference passes $dO$ through the rounding unchanged, so $D$ is still
$\mathrm{rowsum}(P\circ dP)$, the value that $\mathrm{rowsum}(dO\circ O)$ takes at the original output. Reading a copy
$O+E$ instead adds an error term,
\begin{equation}
\mathrm{rowsum}\big(dO\circ(O+E)\big)=D+\mathrm{rowsum}(dO\circ E),
\label{eq:dterm}
\end{equation}
which the control in Figure~\ref{fig:pairing} avoids: it computes $D$ from $P$ and reads no copy of $O$.

Whether the added term averages to zero depends on where the copy came from. Stochastic rounding sends a value bracketed by two
representable numbers up or down at random to one of them, so that its error averages to zero over the random draw
\citep{gupta2015deep}. For such values, a new random rounding uses its own random numbers, so with the forward and $dO$ held fixed
its error still averages to zero, and so does the added term: $D$ is right on average. Reusing the forward's own rounding leaves
nothing to average with the
forward held fixed, and averaging over that rounding need not remove the term either, because the loss saw the rounding
and $dO$ can depend on its error. \citet{qiu2025why} observe such a dependence in BF16 FlashAttention, where $D$ also
reads the forward's own output.

We first ask whether this difference alone can decide the outcome of training when both copies come from the same
distribution. We take FOG's 390M model \citep{hernndezcano2025fully}, trained with the FP8 attention of NVIDIA's
Transformer Engine (TE), at step 999 of two runs (seeds 0 and 1), and fork each checkpoint into three runs that differ
only in what $D$ reads: the forward's rounding, a new rounding, or no copy of $O$. TE rounds its attention output to the
nearest representable value, which gives no second, independently drawn copy, so in these forks the next layer instead
consumes a stochastic rounding of a recomputed $O$ onto TE's FP8 values; a new rounding draws separate random numbers the same way, so the two copies have
the same distribution by construction. We emulate TE's FP8 attention backward in PyTorch so that $D$ can read either
copy or be computed from $P$ (Appendix~\ref{sec:app_setup}), and each run continues for 3000 steps. Forking seed 0's checkpoint again
with a different rounding seed gives a third pair comparing reuse with a new rounding. Each run was
judged by a rule fixed before its results were examined
(Appendix~\ref{sec:app_setup}): it fails if its training loss becomes non-finite or rises more than 1.5 nats above its
lowest value so far, and it trains usefully if it does not fail, its attention logits stay below $10^3$ late in the
run, and its validation loss and late training loss end at most 0.05 nats above the control's.

In all three pairs, reuse fails and the new rounding trains usefully (Figure~\ref{fig:pairing}a: final training losses
\NSOneSevenSixCRZeroSrOneFinalTrain{} and \NSOneSevenSixCRZeroSrZeroFinalTrain{} in the first pair, with the control at
\NSOneSevenSixCRZeroSrBFinalTrain{}). Because FP8 numerics can change between
software releases, we repeated the first pair on PyTorch 2.13 and TE 2.13 (the runs above used 2.11 and 2.10): reuse
again fails and the new rounding trains usefully (Figure~\ref{fig:pairing}b).
So under this controlled stochastic rounding, specifying the distribution of the rounded output does not specify the
learning algorithm.

\subsection{Changing only the backward removes a failure of shipped FP8 attention}\label{sec:outcome-backward}

\begin{figure}[t]
\centering
\includegraphics[width=\linewidth]{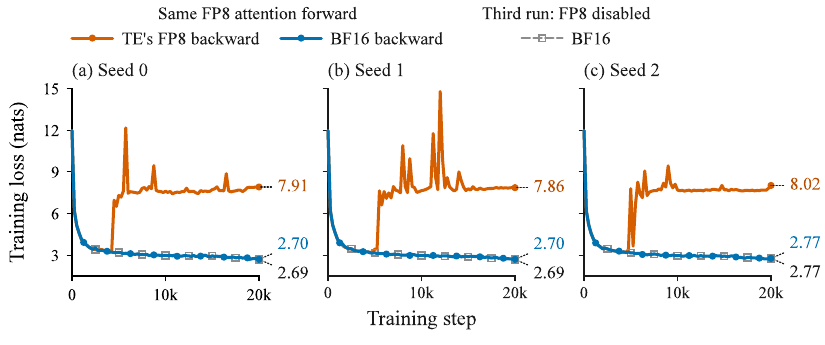}
\caption{\textbf{Changing only the attention backward removes the failure.} Training loss of 390M models trained from
scratch over the 20k-step schedule, one panel per seed, shown every 250th step. With TE's FP8 attention the loss jumps
and stays high on every seed, ending \NSOneSevenFiveSZeroNativeDVal{}, \NSOneSevenFiveSOneNativeDVal{} and
\NSOneSevenFiveSTwoNativeDVal{} nats above BF16 (the same model trained with FP8 disabled) in validation loss. Keeping
its FP8 forward and replacing only the attention backward with a BF16 one tracks BF16, ending
\NSOneSevenFiveSZeroConsDValDisp{}, \NSOneSevenFiveSOneConsDValDisp{} and \NSOneSevenFiveSTwoConsDValDisp{} nats above it.
Seed 2's BF16 run used PyTorch 2.13 and TE 2.13 instead of 2.11 and 2.10 (Appendix~\ref{sec:app_evidence}).}
\label{fig:backward-only}
\end{figure}

We next turn to shipped code: TE's own FP8 attention, with its own backward and its rounding to nearest, trained from
scratch in the setting where FOG reported FP8 attention to fail \citep{hernndezcano2025fully}. We use FOG's 390M model
on PyTorch 2.11 and TE 2.10, with a 20k-step version of FOG's learning rate schedule. The rules are those of
the forks, except that a loss rise on the 100-step grid counts as a failure only if it does not recover by the end of
the run, and both losses must end at most 0.02 nats above those of the same seed trained with FP8 disabled (BF16);
Appendix~\ref{sec:app_setup} gives both rules in full. By these rules TE's FP8 attention fails on all
\NSOneSevenFiveNNativeFails{} seeds we ran: the loss jumps by several nats and stays high (Figure~\ref{fig:backward-only}).

FOG avoided the failure by changing the architecture. Before asking what TE's backward reads, we ask whether changing
the backward alone is enough: we keep the architecture and TE's FP8 forward and replace only the attention backward
with a BF16 one. Our backward calls FlashAttention-2 to recompute attention in BF16 from the forward's FP8 inputs and to compute
the gradients from that recomputation \citep{dao2023flashattention2}, so at $dV$ it uses the recomputed probabilities,
where our reference uses the probabilities TE's forward rounded to FP8 and multiplied by $V$. All three seeds then train usefully
(Figure~\ref{fig:backward-only}). Because scale and software release could each change
the outcome, we repeated the comparison at 1.5B on one seed, each run continued from its own checkpoint at step 2999 to
step 4999, and at 390M on PyTorch 2.13 and TE 2.13 on both seeds we ran, against BF16 on the same software: in each,
TE's FP8 attention fails and the BF16 backward trains usefully, judged by the rule fixed for that setting
(Appendix~\ref{sec:app_evidence}).

\subsection{Replacing only the saved output also removes the failure}\label{sec:outcome-output}

A BF16 backward changes both the precision of the backward and the values it reads. To see whether changing only what
the backward receives is enough, we keep TE's FP8 backward in the original 390M runs from scratch and, through a wrapper
around its call, replace only the saved output it receives, the forward's own output rounded to nearest, with a
stochastic rounding of a BF16 recomputation of $O$ onto the same FP8 values. All three seeds again train usefully,
ending \NSOneEightFourPPoSZeroDValDisp{}, \NSOneEightFourPPoSOneDValDisp{} and \NSOneEightFourPPoSTwoDValDisp{} nats
above the BF16 runs of Figure~\ref{fig:backward-only} in validation loss, while the unchanged runs fail
(Appendix~\ref{sec:app_evidence}). Unlike the forks, this
replacement changes both the source output and its rounding. It shows that what TE's backward receives can decide the
outcome, and that avoiding the failure does not require a more precise output format.

A further fork shows that a copy with larger error at the fork than the forward's own can train where the forward's
own output fails, in TE's backward and in our emulated one
(Appendix~\ref{sec:app_evidence}).

Together, the forks and the runs of TE's own attention show what is at stake: what the backward reads can decide
whether training fails. Which value each use should read is a separate question, and Section~\ref{sec:uses} answers it
with the reference: one tensor can need different values at different uses.

%% file: sections/03-uses.tex
%
\section{One tensor, two uses}\label{sec:uses}

\subsection{The original matches the reference at only one use}\label{sec:uses-norm}

The most accurate value a use could read is the original, yet a normalization output rounded in the forward shows that
even the original cannot serve every use. In each block of our model, the root-mean-square normalization before the
multilayer perceptron (MLP) scales its normalized input $z$ by a trainable gain $\gamma$, giving $u=\gamma\circ z$ \citep{zhang2019root}. Suppose the forward
rounds $u$ to a few bits by stochastic rounding, $u_q=u+r$, the MLP's first layer multiplies the rounded value,
$y=u_qW$, and $u_q$ is saved for the backward; we call this saved value a store. The layer's weight gradient needs the
value the layer multiplied, and the gain's gradient needs the unrounded $z$ that the gain scaled, since the reference
passes the incoming gradient $du=dy\,W^\top$ through the rounding unchanged:
\begin{equation}
dW=u_q^\top dy,\qquad d\gamma={\textstyle\sum_{\mathrm{tokens}}}\, z\circ du .
\label{eq:norm-uses}
\end{equation}
Reading the original at the weight use therefore adds $-r^\top dy$ to $dW$, and reading the rounded value at the gain
use, as $u_q/\gamma$ in place of $z$ (for nonzero $\gamma$), adds $\sum_{\mathrm{tokens}}(r/\gamma)\circ du$ to
$d\gamma$.

\begin{figure}[t]
\centering
\includegraphics[width=\linewidth]{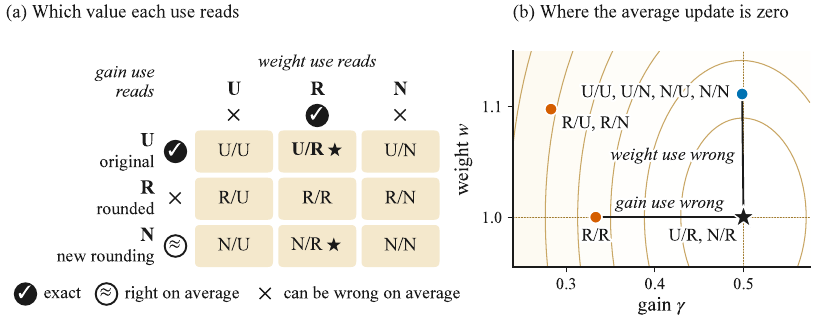}
\caption{\textbf{Each use has its own reference value.} (a) Policies, written gain use/weight use; the marks
rate each use's gradient ($\approx$: on average over the new rounding, the forward fixed; N/N hands both uses one new
rounding). U/R is the reference; $\bigstar$: matches it at both uses, at least on average. (b) Where each policy's
average update is zero in the two-parameter model of Section~\ref{sec:uses-attn}: $\bigstar$ marks the minimum of the expected loss (shading); vermillion dots: the gain use reads R.}
\label{fig:policies}
\end{figure}

Since the original is the average of $u_q$ given $u$, reading it might still seem right on average. It would be, had the
forward multiplied $u$ itself, as in activation-compressed training. Here the loss saw $r$, so $dy$ can depend on it. Even averaged over the forward's rounding, a
looser test than holding the forward fixed, the original's error is then $-\mathbb{E}[r^\top dy]$, which need not
vanish. The same holds for the
rounded value and $du$ at the gain use. A new rounding, averaged over its own draw with the forward held fixed, is right
at the gain use but carries the original's error at the weight use. None of the three values is therefore guaranteed
to be right at both uses, while an assignment by use can be: the reference's, the original at the gain use and the rounded
value at the weight use, or a new rounding at the gain use instead, which is right on average
(Figure~\ref{fig:policies}a).

Whether the original's error at the weight use is detectable in trained models is an empirical question. We take two
390M models trained in BF16 without any store, from two seeds, so that their weights have not adapted to one, and at
probe time insert a 6-bit store at this normalization in all 16 layers; nothing is trained. In all \NSTwoTenCSZeroPThreeCount{} layers of the first model and
all \NSTwoTenCSTwoPThreeCount{} of the second, a detection statistic for the original's average error at the weight use
exceeds the largest value that a control with zero mean reaches in any layer (Appendix~\ref{sec:app_setup}), while at
the gain use its error is exactly zero.

\subsection{No single substitute serves both uses}\label{sec:uses-attn}

A value shared by both uses might still leave the optimum of training in place. It does not in a two-parameter
model of the store whose average updates can be computed exactly. A gain
$\gamma\in(0,1]$ scales an input of one, and the forward consumes a rounding $q\in\{0,1\}$ of $\gamma$ with $P(q=1)=\gamma$,
which is right on average; a weight $w$ multiplies $q$ under a regularized quadratic loss whose expectation has a unique
minimum, at $\gamma=\NSTwoThreeZeroOptOptGamma{}$ and $w=\NSTwoThreeZeroOptOptW{}$ (Appendix~\ref{sec:app_theory}).
Because $q$'s distribution depends on $\gamma$, the gradient of the expected loss has one more term, which every policy
receives alike. Reading $\gamma$ at the gain use and $q$ at the weight use is then right on average everywhere, as is a
new draw at the gain use, so both leave the minimum stationary. None of the three values read by both uses does
(Figure~\ref{fig:policies}b): reading $q$ at both, for instance, is stationary at $\gamma=\NSTwoThreeZeroOptRRGamma{}$
and $w=\NSTwoThreeZeroOptOptW{}$, where the expected loss exceeds its minimum by
\NSTwoThreeZeroOptRRExcess{}.

For attention, the operation of Section~\ref{sec:outcome-backward}, a theorem rules out every shared substitute for
its rounded probabilities. FP8 attention rounds the probabilities $P$ before multiplying them by $V$, so the rounded
probabilities
$p_q=p+r$ of a row have two uses: the gradient of $V$ needs the rounded values that multiplied $V$, while the softmax's
backward needs its Jacobian at the probabilities the softmax produced, $J(p)=\mathrm{diag}(p)-pp^\top$. Suppose one
row's rounding is right on average but not exact, independent across entries, and
confined for each entry to a fixed pair of neighbouring representable values, and the loss is quadratic in the row's
output. A shared substitute $Z$ is used as $(Z,J(Z))$ at the gradient of $V$ and the softmax's backward, respectively.
No such substitute matches both reference gradients on average with the forward held fixed, for every incoming
gradient at each use. Being right at the gradient of $V$ forces $Z$ to vary
across the forward's roundings at least as much as $p_q$, and the Jacobian, quadratic in the probabilities, turns that
variation into an error. Appendix~\ref{sec:app_theory} gives the proof, and an exact enumeration in two cases
that confirms the predicted errors of four assignments.

An assignment by use remains possible: the reference reads $p_q$ at the gradient of $V$ and $p$ in the softmax's
backward.

\subsection{Comparable final loss does not show that a policy follows the reference}\label{sec:uses-loss}

The remaining check, final loss, needs models trained with the normalization store. We trained the 390M model in BF16
with the store at the normalization before each MLP as its only low-precision rounding, for
8000 steps on two seeds: seven policies at 6 bits, which share the forward and differ only in what the backward reads,
and five at 5 bits (Appendix~\ref{sec:app_setup}). Besides the policies of Figure~\ref{fig:policies}a, both widths ran
U/R+$\varepsilon$, a control for noise whose weight use reads the rounded value plus a new rounding's residual. Before
reading each store's final losses we named the comparisons to read, and before reading any we fixed a margin of
$\pm$\NSTwoOneOneDeltaNats{} nats: two policies count as equivalent when the model-based 95\% interval of their
difference in held-out loss, evaluated without the store, lies inside it.

\begin{figure}[t]
\centering
\includegraphics[width=\linewidth]{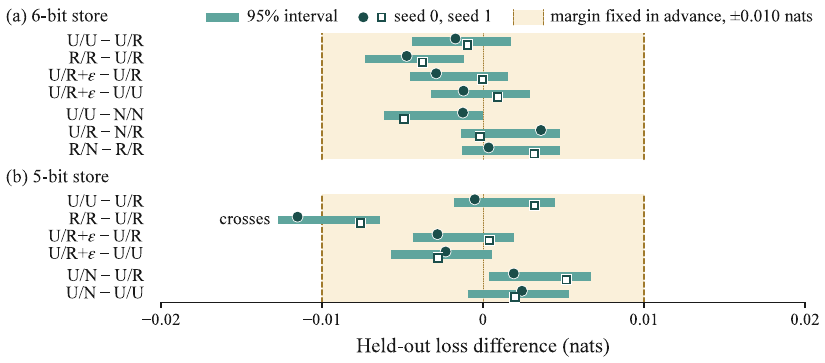}
\caption{\textbf{Every named interval but one lies inside the margin fixed in advance.} Held-out loss of
the 390M runs trained with a 6-bit (a) or 5-bit (b) store, evaluated without it, first policy minus second: model-based
95\% intervals (bars) and the two seeds (markers). Only the R/R$\,-\,$U/R interval at 5 bits crosses the margin.}
\label{fig:endpoint}
\end{figure}

At 6 bits every named interval lies inside the margin, including that of the original at both uses (U/U) against the
reference assignment (U/R) (Figure~\ref{fig:endpoint}a). The rounded value at both uses (R/R) ends
\NSTwoOneOneKSevenDBar{} nats from U/R: lower, with an interval that excludes zero, yet inside the margin. At 5 bits
R/R ends \NSTwoOneFiveDoseFiveKSixDBar{} nats from U/R, and its interval, which excludes zero, crosses the margin's
lower edge; every other named interval stays inside (Figure~\ref{fig:endpoint}b).

\begin{figure}[t]
\centering
\includegraphics[width=\linewidth]{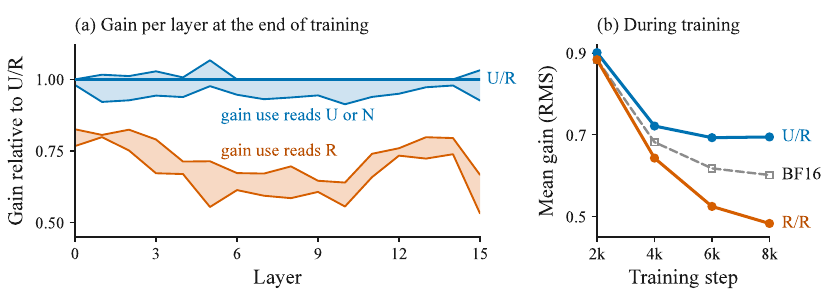}
\caption{\textbf{Comparable final loss, different trained weights.} (a) Root-mean-square gain of the normalization before
each MLP at the end of the 6-bit runs, divided by U/R's at the same layer and seed; each band spans its group's runs on
both seeds (U/R's at 1). (b) The mean gain over layers during training for U/R, R/R and BF16 without the store (seed 0).}
\label{fig:gains}
\end{figure}

Nor do comparable losses mean the same trained model. The gain's gradient is the store's other use, and in the 6-bit
runs the two policies whose gain use reads the rounded value end with a lower root-mean-square gain than the five
others in every layer, on both seeds (Figure~\ref{fig:gains}).

The error itself could also change as the model adapts
to the store. To see how, we continued two runs trained without the store, one per seed, for 750 steps with the 6-bit store
under U/R, U/U or R/R, or with a forward that does not consume the store, as a control, and probed every continuation
with the same store (Appendix~\ref{sec:app_setup}). Under U/R and U/U, the squared norm of the average error that reading
the original would introduce at the last layer's weight use shrank relative to the control, yet the error stayed detectably
nonzero on both seeds; under R/R, the root-mean-square gain fell below U/R's in every layer
by the first probe after the switch.

A policy that departs from the reference thus runs a different learning algorithm, which its final loss, comparable to
or lower than the reference's, does not reveal.
Whether a policy follows the reference has to be checked at each use, as Section~\ref{sec:contract} does at the
operator.

%% file: sections/04-contract.tex
%
\section{Checking each use against the reference}\label{sec:contract}

\subsection{Each use has a checkable requirement}\label{sec:contract-req}

Final loss judges a whole run, whereas the reference makes a statement about each use, so a policy can be checked one
use at a time. Because the reference passes gradients through each rounding unchanged, it takes each use's gradient at
the actual inputs and outputs of its forward operation. The weight use of Section~\ref{sec:uses-norm} and the gradient of
$V$ therefore need the rounded value. The gain use, the correction term $D$ and the softmax's backward need the original.
A use may instead read a substitute, such as a new rounding, if averaging over the substitute's own draw gives the
reference gradient conditional on the forward, including its rounding, and the incoming gradient it actually receives.
This conditioning accounts for any random dependence between the substitute and that gradient. Two uses can share one value
only if it meets both requirements, and at the normalization output of
Section~\ref{sec:uses} no value is guaranteed to.

Each requirement concerns one use of one operator, so checking it needs no training run. Our checker takes a policy written down use
by use, together with the random draws on which each incoming gradient and each rule's gap from the reference depend. It
reports whether each condition of each requirement holds, fails, or cannot be decided. We test its verdicts by exact computation on small finite cases
and on deliberately broken policies (Appendix~\ref{sec:app_contract}). At the store of Section~\ref{sec:uses}, it
certifies a weight use that reads the rounded value its layer multiplied. For a gain use that reads that rounding too,
it reports the requirement as undecided: the incoming gradient depends on the same random draw, so the error need not
average away.

\subsection{The requirements predict when reuse matters}\label{sec:contract-predict}

The dependence that left the gain use undecided also yields predictions about existing code, which need not declare
any policy.
Consider the contrast of Figure~\ref{fig:pairing}: a use that needs the original reads either the forward's own
rounding, as under reuse, or a new rounding from the same distribution. The two copies are equally accurate and differ
in one way: the incoming gradient can depend on the forward's rounding but not on the new one
(Section~\ref{sec:outcome-forks}). Averaged over the forward's rounding as well as the new one, reuse can therefore change
what the backward computes only if the incoming gradient depends on the forward's rounding. Whether it does is decided by
the computation around the use, such as what is recomputed and the loss.

\begin{figure}[t]
\centering
\includegraphics[width=\linewidth]{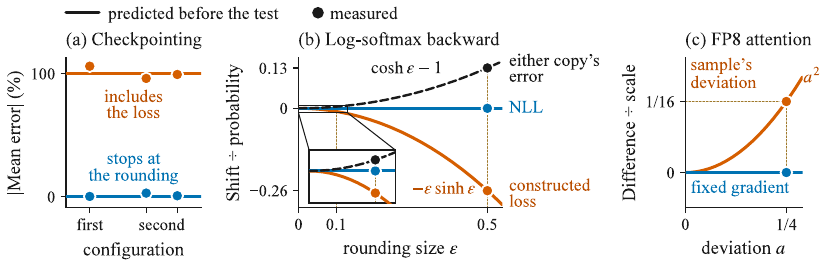}
\caption{\textbf{Registered predictions about reuse held.} Lines: predictions written before each test; dots:
measurements. (a)~The use's mean error, as a percentage of its size without checkpointing (random state not
restored). (b)~Mean change per gradient entry, divided by its probability: shared minus
independent copy under each loss, and either copy's error under the likelihood; the inset enlarges
$\varepsilon=0.1$. (c)~Query gradient, independent minus shared sample, divided by the softmax scale; mean of four
sample pairs.}
\label{fig:prospective}
\end{figure}

We tested such predictions on three pieces of code that we did not write: PyTorch's checkpointing and log-softmax
backward, and TE's FP8 attention backward. Before measuring, we wrote
down each prediction and how its result would be read. Each test isolates one
backward operator: our code sets the rounding or sampling, and nothing is trained. Figure~\ref{fig:prospective} summarizes these predictions and their
measurements.

\subsection{Predictions about reuse held on three operators}\label{sec:contract-tests}

Checkpointing varies what is recomputed. By default it restores the random state before recomputing, so the
recomputation reproduces the forward's random numbers \citep{pytorch2026checkpoint}. A dropout mask needs this, since the
forward multiplied it; a PyTorch issue reports training that collapsed when a compiler's recomputation drew
new masks \citep{pytorch2026rngrecompute}. Our two-layer test model, run eagerly on CPU in PyTorch 2.14, has both a dropout mask
and, with our own stochastic rounding, a use that needs the original but reads the recomputed rounding. We predicted
that restoring the random state reproduces that use's recorded error bit for bit. Without restoration the mask changes.
Whether the effect of reuse also vanishes depends on the incoming gradient. If recomputation stops at the rounding, the
incoming gradient still comes from the original draw and the effect should vanish; if recomputation includes the loss,
the incoming gradient is recomputed from the new draw and the effect should stay. Both
configurations we registered agreed. In the first, without restoration, the use's mean error shrank to
\NSOneNineOneOutFPctOfBase\,\% of its size without checkpointing when recomputation stopped at the rounding, and stayed
at \NSOneNineOneInFPctOfBase\,\% when it included the loss
(Figure~\ref{fig:prospective}a). One switch restores the
random state for the whole recomputed part, so in our model, when that part ends at the rounding, neither setting serves
both the mask and the use.

The log-softmax test varies the loss instead. PyTorch's CPU backward for log-softmax exponentiates its saved output,
into which our code injects a rounding error of $\pm\varepsilon$. Under negative log-likelihood with a fixed target, the
incoming gradient does not depend on that output, and we predicted that reuse leaves the average gradient unchanged.
Under a second loss, constructed by our code so that the incoming gradient equals the rounding error, we predicted that
reuse shifts each gradient entry by $-\varepsilon\sinh\varepsilon$ times its probability on average. Both predictions
held at both values of $\varepsilon$, 0.1 and \NSTwoTwoSixLSEps{}
(Figure~\ref{fig:prospective}b). Yet the two copies agreeing does not make
either right: under the likelihood, the gradient from either copy is off on average by $\cosh\varepsilon-1$ times each
entry's probability, as predicted, so here not even a new rounding is right.

After two tests on CPU, the last takes the contrast of Figure~\ref{fig:pairing} to a shipped fused kernel: TE's FP8
attention backward, whose saved output Section~\ref{sec:outcome-output} replaced. We tested it in TE 2.13 with cuDNN
9.20, on one attention row in which every value that decides the result is exactly representable in FP8. Our code moves one
entry of the saved output up or down by $a$ from its mean. The backward then reads either an independent sample or, as
under reuse, the sample that the incoming gradient depends on. For a backward that computes $D$ from the saved output,
as FlashAttention's does \citep{dao2022flashattention}, we predicted, for each of the four sample pairs, the change in
the query gradient at that entry from the shared sample to the independent one. On average, the change should
be $a^2$ times the softmax scale when the incoming gradient is the sample's deviation from its mean, and zero when the
incoming gradient is fixed. All eight predicted values were reproduced exactly (Figure~\ref{fig:prospective}c shows their
means).

Every prediction about reuse held, including the predictions that reuse changes nothing;
Appendix~\ref{sec:app_contract} reports every registered statement with its outcome. The requirements thus give a rule
that is stated for each use and checked on its operator before any training run. That appendix's two-stage example shows why
the conditioning matters. Both stages match the reference on average at a fixed input, but sharing a random draw biases
their composition. Conditioning on the received input detects this error.

%% file: sections/05-conclusion.tex
%
\section{Conclusion}\label{sec:conclusion}

Low-precision training leaves the backward a choice wherever it reads a rounded tensor again: the rounded value the
forward used, the original, or a new rounding. We have argued that this choice is part of the learning algorithm, not a
detail of memory and precision. For attention's correction term, two copies from the same rounding distribution led
training to opposite outcomes, and changing only the backward removed a failure of shipped FP8 attention. Under the
declared reference, one stored tensor can need different values at different uses, so even the exact original departs
from it at one of them. Trained models carry that discrepancy while their final loss stays comparable. Copy
accuracy and final loss therefore cannot settle the choice; a requirement at each use can, and it is checked on its
operator without training. In three tests on PyTorch and Transformer Engine operators, every prediction the
requirements made about reuse held.

The choice arises wherever a backward reads forward state that was saved, compressed or recomputed. An implementation
should make it explicitly, stating which value each use reads and testing whether the use's incoming gradient depends
on the rounding it reads. What the backward reads
decides which gradient training follows, and so belongs to the learning algorithm.

%% file: sections/A-related.tex
%
\section{Related work}\label{sec:app_related}

\paragraph{Low-precision training and attention.}
Recipes for FP8 and 4-bit training choose formats and scalings
\citep{peng2023training,blake2023unit,xi2024coat,fishman2024scaling,mishra2025recipes}, and some also choose
rounding rules \citep{tseng2025training,castro2025quartet,panferov2026quartet}. \citet{chmiel2025all} choose rounding
separately at each of six quantization points and keep an FP4 forward with a BF16 backward during a short fine-tuning
phase. SwitchBack computes weight gradients at higher precision than input gradients, and its SwitchBackM variant
saves 8-bit copies for the backward \citep{wortsman2023stable}. \citet{kumar2024scaling} study precision scaling laws with
only the forward quantized. For attention, FOG changes
the architecture to avoid divergence \citep{hernndezcano2025fully}, while low-bit kernels choose both the products run in
low precision and the forward values saved or recomputed for the backward
\citep{zhang2025sageattention,zhang2026sagebwd,hu2026hardware}. \citet{qiu2025why} trace a BF16 FlashAttention failure to the correction term and attribute its error to biased
rounding in the attention output. Other studies predict loss explosions from spectral measurements
\citep{qiu2025spectral}, reproduce them in small models \citep{wortsman2023small}, or measure numerical deviation
\citep{golden2024flash}. \citet{xie2026one} connect low-precision errors from several sources to runaway query and key
projections. These works motivate the controlled comparison in Section~\ref{sec:outcome-forks}: keeping the rounding
distribution fixed while changing which copy the backward reads.

\paragraph{Separate copies and activation compression.}
Attn-QAT keeps a high-precision attention output for the backward \citep{zhang2026attn}. With one-dimensional block
scaling, including MXFP8 \citep{mishra2025recipes}, Transformer Engine quantizes rowwise and columnwise copies separately
from the unrounded tensor and notes that their differences may affect gradients \citep{nvidia2026blockwise}.
\citet{rahimifar2026stable} preserve FP4 block scales under transposition by using square blocks for weights and gradients.
Our requirements concern which value each use needs under a declared gradient reference.
PyTorch's hard Gumbel-softmax uses a one-hot sample in the forward and the soft relaxation of the same draw for its
gradient \citep{jang2017categorical,pytorch2026gumbelsoftmax}. Activation-compressed training instead leaves the forward
exact and compresses what the backward reads. ActNN shows that two independently quantized copies of a normalization
input give an unbiased gradient \citep{chen2021actnn}; GACT shows that an unbiased compressor can still bias gradients
\citep{liu2022gact}. PRAC gives an unbiased weight gradient for one linear layer from an unbiased reconstruction
\citep{li2026prac}, where the reconstruction has one use and the incoming gradient is independent of its draw.
Section~\ref{sec:uses-norm} addresses a different setting: the forward consumes the rounded value, and one tensor feeds
two gradients that require different values, even when the original is available.

\paragraph{Gradient estimators.}
Our reference uses a straight-through rule \citep{bengio2013estimating,yin2019understanding}; the backward-state policy
chooses which values each use reads within that rule. \citet{weber2019credit} state when a factor inside
backpropagation may be replaced by its conditional expectation. Once the loss has seen the rounding, that condition
need not hold. Related sampling choices occur in inverse rendering, where estimates of the primal image and its derivative are
sampled separately to avoid bias \citep{azinovi2019inverse,vicini2021path}, and in Flipout, which decorrelates gradients
across examples \citep{wen2018flipout}. \citet{li2026understanding} argue for deliberate bias in straight-through
quantization-aware training, whose gradients are evaluated at the deployed quantized weights. Our question is whether
a policy follows its declared reference; a lower final loss does not establish this (Section~\ref{sec:uses-loss}).

\paragraph{Recomputation and saved state.}
Checkpointing restores random state to replay values such as dropout masks that the forward multiplied
\citep{pytorch2026checkpoint}. A PyTorch issue reports the effect of recomputing dropout with new seeds and proposes
saving the forward random outputs \citep{pytorch2026rngrecompute}. This replay requirement differs from the requirement
for a use that needs the original but reads a rounding, as the checkpointing test in
Section~\ref{sec:contract-tests} shows. QEffect specifies runtime contracts for FP8 pipeline training, covering updates
to scaling factors, ownership of backward work, versions of cached weights and GPU stream completion
\citep{chen2026explicit}. Our requirements take the derivative formulas as given and specify which value each use reads
(Section~\ref{sec:contract-req}).

%% file: sections/B-theory.tex
%
\section{Theory}\label{sec:app_theory}

Policies are named as in Figure~\ref{fig:policies}a. The first letter is the value read at the use that needs
the original, and the second the value read at the use that needs the rounded value. U is the original, R the rounded
value the forward used, and N a new rounding from the same distribution.

\subsection{No shared value keeps the minimum of the two-parameter model}\label{sec:app_theory-model}

The model reduces the normalization store to one token and one channel. A gain $\gamma\in(0,1]$ scales an input of one,
so $\gamma$ is also the value that the forward rounds. The forward consumes a rounding $q\in\{0,1\}$ of $\gamma$ with
$P(q=1)=\gamma$, which is right on average, and a weight $w\in\mathbb{R}$ multiplies $q$. One draw's loss has two
regularizing terms, which place the minimum of the expected loss inside the domain:
\begin{equation}
\ell=\tfrac12(wq-1)^2+2\big(\gamma-\tfrac38\big)^2+(w-1)^2,\qquad
\bar\ell(\gamma,w)=\mathbb{E}_q[\ell]=\tfrac{9}{32}+2\big(\gamma-\tfrac12\big)^2+\big(1+\tfrac{\gamma}{2}\big)(w-1)^2 .
\label{eq:app-model-loss}
\end{equation}
So $\bar\ell$ has a unique minimum, at $\gamma=\NSTwoThreeZeroOptOptGamma{}$ and $w=\NSTwoThreeZeroOptOptW{}$, where
$\bar\ell=\NSTwoThreeZeroOptJStar{}$.

The backward passes the incoming gradient $\zeta=\partial\ell/\partial q=w(wq-1)$ through the rounding unchanged. A policy
that reads $x_\gamma$ at the gain use and $x_w$ at the weight use updates
\begin{equation}
d\gamma=\zeta\,x_\gamma/\gamma+4\big(\gamma-\tfrac38\big)+c,\qquad dw=(wq-1)\,x_w+2(w-1),\qquad c=w^2\big(\tfrac12-\gamma\big).
\label{eq:app-model-update}
\end{equation}
The gain use divides by $\gamma$ because it needs the input that the gain scaled. The reference reads $x_\gamma=\gamma$ and
$x_w=q$, so on average it collects $\mathbb{E}[\zeta]=w(w\gamma-1)$ through $q$. The gradient of $\bar\ell$ collects
$\ell(1)-\ell(0)=\tfrac12w^2-w$ instead, because $q$'s distribution depends on $\gamma$. Their difference is $c$, the term
that Section~\ref{sec:uses-attn} says every policy receives alike. Adding it to every update makes the reference's average
update exactly $\nabla\bar\ell$, so that policies differ only in the values they read. It reads no stored value and is
zero at the minimum, so stationarity at the minimum does not depend on it; the other rest points below are those of the
updates with $c$.

Averaging~\eqref{eq:app-model-update} over $q$ and over any new rounding $f$ gives each policy's average update as
$\nabla\bar\ell$ plus one error per use. Since $f$ is independent of $q$ with mean $\gamma$, reading U or N at the gain use
adds nothing. Reading R there adds $\mathbb{E}[\zeta(q-\gamma)]/\gamma=w^2(1-\gamma)$, because $\zeta$ depends on $q$.
Reading R at the weight use adds nothing, while reading U or N there adds $-\mathbb{E}[(wq-1)(q-\gamma)]=-w\gamma(1-\gamma)$.
At the minimum, for instance, R/R's average update is $(\NSTwoThreeZeroOptRRGGammaAtOpt{},0)$ and U/U's is
$(0,\NSTwoThreeZeroOptSSGWAtOpt{})$.

With three values at each of two uses, the nine policies fall into four classes with equal average updates, and
Figure~\ref{fig:policies}b marks where each class's average update is zero. For U/R and N/R both errors vanish, so their
average update is $\nabla\bar\ell$ everywhere and is zero only at the minimum. For R/R, the weight component
$(2+\gamma)(w-1)$ vanishes only at $w=1$, and the gain component is then $4(\gamma-\tfrac12)+(1-\gamma)=3\gamma-1$. R/R is
therefore stationary at $\gamma=\NSTwoThreeZeroOptRRGamma{}$ and $w=\NSTwoThreeZeroOptOptW{}$, where $\bar\ell$ exceeds its
minimum by $2(\gamma-\tfrac12)^2=\NSTwoThreeZeroOptRRExcess{}$. For U/U, U/N, N/U and N/N, the weight component vanishes at
$w=(2+\gamma)/(2+\gamma^2)$, and the gain component then vanishes where
\begin{equation}
8\big(\gamma-\tfrac12\big)(\gamma^2+2)^2+\gamma^2(1-\gamma)^2=0 .
\label{eq:app-model-uu}
\end{equation}
R/U and R/N share that value of $w$ and add $2(\gamma+2)^2(1-\gamma)$ to the left side of~\eqref{eq:app-model-uu}. Each of
the two quintics has a positive derivative on $[0,1]$ and changes sign there, so it has exactly one root in $(0,1)$, which
gives its class's point in Figure~\ref{fig:policies}b.

No value read at both uses keeps the minimum stationary, whether U, R, N or any other. At the minimum, $\zeta=q-1$ and
$c=0$, so a value $x$ read at both uses gives $d\gamma=2(q-1)x+\tfrac12$ and $dw=(q-1)x$. Here $x$ may depend on $q$, on a
new rounding and on $\gamma$, as long as $\kappa=\mathbb{E}[(q-1)x]$ is defined. The average update is then
$(2\kappa+\tfrac12,\kappa)$: its weight component vanishes only if $\kappa=0$, and its gain component is then $\tfrac12$.
An assignment by use escapes this because it reads different values at the two uses.

\subsection{No shared substitute serves both uses of attention probabilities}\label{sec:app_theory-thm}

The theorem of Section~\ref{sec:uses-attn} concerns one row of attention. The row has scores $s\in\mathbb{R}^n$,
probabilities $p=\mathrm{softmax}(s)$ and values $V\in\mathbb{R}^{n\times d}$ with rows $v_j^\top$. The forward rounds $p$
to $p_q=p+r$ and outputs $o=V^\top p_q$. We write $F$ for the forward's inputs $s$ and $V$, so that an average given $F$
is also taken over the forward's rounding, while an average given $F$ and $r$ holds the forward fixed. The theorem makes
the assumptions of Section~\ref{sec:uses-attn}. Each entry's rounding moves $p_j$ to one of a fixed pair of neighbouring
representable values. The rounding is right on average, $\mathbb{E}[r\mid F]=0$, and independent across entries. It is not
exact: its covariance $\Sigma_r=\mathrm{diag}(\sigma^2)$ is not zero. The loss $L$ is quadratic in $o$. The independence
and the quadratic loss also give the closed forms of Section~\ref{sec:app_theory-enum}, and the pairs of values make its
enumeration finite.

With $h=\partial L/\partial o$ the row's incoming gradient and $dp=Vh$ the gradient that reaches the probabilities, the
reference computes
\begin{equation}
dV=p_q\,h^\top,\qquad ds=J(p)\,dp,\qquad J(x)=\mathrm{diag}(x)-xx^\top .
\label{eq:app-ref}
\end{equation}
For the program $o=V^\top(\mathrm{softmax}(s)+r)$ with $r$ held fixed, these are the ordinary derivatives. A policy reads
a vector $\hat p$ at the gradient of $V$ and a matrix $\hat J$ at the softmax's backward, so its errors are
\begin{equation}
\Delta_V=(\hat p-p_q)\,h^\top,\qquad \Delta_s=(\hat J-J(p))\,dp .
\label{eq:app-errors}
\end{equation}
We call a policy right on average at a use, with the forward held fixed, for every incoming gradient, if its error there
has zero average given $F$ and $r$ whatever the incoming gradient. The incoming gradient may be any bounded function of $F$
and $r$: $h$ at the gradient of $V$, and $dp$ at the softmax's backward. A shared substitute is one random vector $Z$ with
finite second moments, read at both uses: $\hat p=Z$ and $\hat J=J(Z)$. $Z$ may use random numbers of its own.

\paragraph{Theorem.} \emph{In this setting, no shared substitute is right on average, with the forward held fixed, at
both uses for every incoming gradient.}

\emph{Proof.} Suppose $Z$ is. Taking $h$ and $dp$ to be the standard basis vectors gives
$\mathbb{E}[Z\mid F,r]=p_q$ and $\mathbb{E}[J(Z)\mid F,r]=J(p)$. Averaging the first identity over the forward's rounding
gives $\mathbb{E}[Z\mid F]=p$, and the law of total covariance gives
\begin{equation}
\mathrm{Cov}(Z\mid F)=\mathrm{Cov}(p_q\mid F)+\mathbb{E}\big[\mathrm{Cov}(Z\mid F,r)\,\big|\,F\big]\succeq\Sigma_r ,
\label{eq:app-totalcov}
\end{equation}
where $\succeq$ is the positive semidefinite order. Because $J$ is quadratic, $\mathbb{E}[J(Z)\mid F]=\mathrm{diag}(\mathbb{E}[Z\mid F])-\mathbb{E}[ZZ^\top\mid F]
=J(p)-\mathrm{Cov}(Z\mid F)$. Averaging the second identity over the forward's rounding gives
$\mathbb{E}[J(Z)\mid F]=J(p)$. Hence $\mathrm{Cov}(Z\mid F)=0$, which contradicts~\eqref{eq:app-totalcov} because
$\Sigma_r$ is positive semidefinite and not zero.\hfill$\square$

The proof also sizes the conflict. Take any $Z$ that is right at the gradient of $V$. The average of the matrix that it
hands the softmax's backward falls short of $J(p)$ by $\mathrm{Cov}(Z\mid F)\succeq\Sigma_r$. By~\eqref{eq:app-totalcov},
the shortfall is exactly $\Sigma_r$ only when $Z=p_q$, as under R/R, and any randomness of $Z$'s own adds to it.

\subsection{The errors of four policies, predicted and enumerated exactly}\label{sec:app_theory-enum}

The theorem covers every shared substitute at once; for particular policies the errors can be computed in closed form.
Take $L=\tfrac12\lVert o\rVert^2$, so that $h=o=V^\top p_q$ and $dp=Gp_q$ with $G=VV^\top$. Write $g_j=\lVert v_j\rVert^2$
for $G$'s diagonal, $\tau=\sum_k g_k\sigma_k^2$, and $\mu_3$ for the vector of the rounding's third moments. A new
rounding $p_f=p+r_f$ has the same distribution as $p_q$ and is independent of it given $F$.
Table~\ref{tab:app-theory-errors} lists the average errors given $F$ of the reference and four other policies. A nonzero
average given $F$ rules out a zero average given $F$ and $r$, so each nonzero entry shows that its policy fails the
theorem's condition at that use.

\begin{table}[t]
\centering
\caption{\textbf{Average errors of the reference and four policies at the two uses of an attention row's rounded
probabilities.} Policy codes name the softmax's backward first; averages are over the forward's rounding and any new
rounding, the forward's inputs fixed; $\delta$ is given in~\eqref{eq:app-replay-error}. Zeros marked * hold in every draw.}
\label{tab:app-theory-errors}
\small
\begin{tabular}{@{}lcccc@{}}
\toprule
Policy & Softmax's backward reads & Gradient of $V$ reads & $\mathbb{E}[\Delta_s\mid F]$ & Row $j$ of $\mathbb{E}[\Delta_V\mid F]$ \\
\midrule
U/R (reference) & $J(p)$ & $p_q$ & $0^*$ & $0^*$ \\
R/R & $J(p_q)$ & $p_q$ & $\delta$ & $0^*$ \\
U/U & $J(p)$ & $p$ & $0^*$ & $-\sigma_j^2\,v_j^\top$ \\
N/N & $J(p_f)$ & $p_f$ & $-\Sigma_rGp$ & $-\sigma_j^2\,v_j^\top$ \\
R/U & $J(p_q)$ & $p$ & $\delta$ & $-\sigma_j^2\,v_j^\top$ \\
\bottomrule
\end{tabular}
\end{table}

Each entry keeps the terms that survive zero mean and independence across entries. At the gradient of $V$,
$\mathbb{E}[r_jh^\top\mid F]=\mathbb{E}[r_jr^\top V\mid F]=\sigma_j^2v_j^\top$ gives the three nonzero entries, and a
new rounding contributes nothing because $h$ does not depend on it. At the softmax's backward, R/R's error is
$(J(p+r)-J(p))\,G(p+r)$, with $J(p+r)-J(p)=\mathrm{diag}(r)-pr^\top-rp^\top-rr^\top$. Of the eight products, the three
that are linear in $r$ average to zero, and the other five give
\begin{equation}
\delta=\underbrace{g\circ\sigma^2}_{\mathrm{diag}(r)Gr}\;\underbrace{-\,\tau p}_{-pr^\top Gr}\;
\underbrace{-\,2\Sigma_rGp}_{-rp^\top Gr\,-\,rr^\top Gp}\;\underbrace{-\,g\circ\mu_3}_{-rr^\top Gr} .
\label{eq:app-replay-error}
\end{equation}
N/N's new rounding is independent of $dp$, and $\mathbb{E}[J(p_f)\mid F]=J(p)-\Sigma_r$, so its error averages to
$-\Sigma_rGp$.

We computed these averages exactly, with rational arithmetic, over every rounding outcome, and formed each backward from
its operands rather than from the closed forms. Cell A has $n=4$, $d=3$, $p=(\tfrac25,\tfrac14,\tfrac15,\tfrac3{20})$ and
value rows $(1,0,2)$, $(0,1,-1)$, $(2,-1,1)$ and $(3,1,0)$; cell B has $n=3$, $d=2$, $p=(\tfrac12,\tfrac3{10},\tfrac15)$
and value rows $(1,1)$, $(2,-1)$ and $(0,3)$. Each cell used two rounding laws. The first is stochastic rounding to a
grid of spacing $\tfrac18$ in cell A and $\tfrac16$ in cell B, whose third moments are not all zero. Under it, one entry
of each cell lies on the grid and is never rounded. The second is the symmetric law $r_j=\pm\tfrac18$, each sign with
probability $\tfrac12$, whose third moments are zero.

All eleven statements fixed before the enumeration code was written resolved as predicted: the ten true statements
held, and the deliberately false statement that N/N has no error at either use was refuted. The controls checked zero
rounding and the third-moment term $g\circ\mu_3$. Zeros were checked both per draw and on average. Six deliberate errors planted in the code changed their intended outcomes.

N/N can look like a repair to a test that looks only at the softmax's backward. In $L_1$ norm its error there is
\NSOneNineThreeASrRefreshOverReplay{} of R/R's in cell A under stochastic rounding and
\NSOneNineThreeASymRefreshOverReplay{} under the symmetric law, and \NSOneNineThreeBSrRefreshOverReplay{} and
\NSOneNineThreeBSymRefreshOverReplay{} in cell B. The new rounding removes the terms that pair the rounding with $dp$, but
it turns R/R's exact zero at the gradient of $V$ into $-\sigma_j^2v_j^\top$.

\subsection{One new rounding suffices at the softmax's backward}\label{sec:app_theory-copy}

The theorem excludes one value read at both uses, not a second rounding. What fails is plugging one new rounding into
$J$: for a new rounding $a$, $\mathbb{E}[J(a)\mid F]=J(p)-\Sigma_r$, because the diagonal of $aa^\top$ holds the squares
$a_i^2$. With $\mathbf 1$ the vector of ones, the operator
\begin{equation}
K(a)=\mathrm{diag}\big(a\,(\mathbf 1^\top a)\big)-aa^\top
\label{eq:app-onecopy}
\end{equation}
avoids those squares: its diagonal entries are $a_i\sum_{k\neq i}a_k$ and its off-diagonal entries $-a_ia_j$, products
of different entries. Assume that $a$'s entries are independent given $F$, that $\mathbb{E}[a\mid F]=p$, and that $a$
is independent of the forward's rounding given $F$. Then the off-diagonal entries average to $-p_ip_j$ and the diagonal
entries to $p_i(1-p_i)$, because $\mathbf 1^\top p=1$. So $\mathbb{E}[K(a)\mid F,r]=J(p)$, and every row of $K(a)$ sums to
zero in every draw, as the rows of $J(p)$ do. Reading $p_q$ at the gradient of $V$ and $K(a)$ at the softmax's backward is
therefore right on average at both uses. It needs a rounding that the forward did not consume.

The eleven statements covered three operators built from new roundings. For two independent new roundings $a$ and $b$,
$J_2(a,b)=\mathrm{diag}\big(\tfrac{a+b}2\big)-\tfrac12(ab^\top+ba^\top)$ averages to $J(p)$, so its error is zero on
average. Plugging one new rounding into $J$ leaves $-\Sigma_rGp$. Using the forward's own rounding as one of $J_2$'s two
roundings leaves $\tfrac12(g\circ\sigma^2-\tau p-\Sigma_rGp)$, because that rounding is paired with $dp$. We subsequently checked $K$ by the same enumeration in cell A.

\subsection{What R/R adds to the reference's update is not a gradient}\label{sec:app_theory-curl}

Since R/R departs from the reference, one may ask whether it follows the gradient of some other objective. In the two
cells of Section~\ref{sec:app_theory-enum}, under two symmetric laws, it does not. Write $\theta=(s,V)$, take
$L=\tfrac12\lVert o\rVert^2$ as in Section~\ref{sec:app_theory-enum}, and let $m_{\mathrm{P}}(\theta)$ be policy P's
average update given $F$. For the reference, $m_{\mathrm{U/R}}=\big(J(p)Gp,\,(pp^\top+\Sigma_r)V\big)$. The average loss
$\bar L(\theta)=\mathbb{E}[L\mid F]=\tfrac12p^\top Gp+\tfrac12\sum_j\sigma_j^2g_j$ has the gradient $m_{\mathrm{U/R}}+c$,
with $c=\big(\tfrac12J(p)((\sigma^2)'\circ g),\,0\big)$ and $(\sigma^2)'_j=d\sigma_j^2/dp_j$. As in
Section~\ref{sec:app_theory-model}, $c$ comes from the rounding's law depending on $p$, and it is zero when the law does
not. Derivatives are taken on an open set where each $\sigma_j^2$ is a differentiable function of $p_j$, as under the
symmetric laws below. Under stochastic rounding this holds only inside a grid cell, and cells A and B each put one entry on
a grid point.

R/R reads the same value as U/R at the gradient of $V$, so their average updates differ only in the scores, by R/R's
error $\delta$ at the softmax's backward. The difference is the one-form
\begin{equation}
\beta=(m_{\mathrm{R/R}}-m_{\mathrm{U/R}})\cdot d\theta=\textstyle\sum_i\delta_i\,ds_i .
\label{eq:app-oneform}
\end{equation}
With $c$ added to both, R/R's average update is $\nabla\bar L+(\delta,0)$. This update is the gradient of a function near a
state only if $\beta$ is the differential of a function there, which requires $d\beta=0$, that is, a symmetric Jacobian of
$(\delta,0)$. On a simply connected domain where $d\beta=0$ throughout, the converse also holds (the Poincar\'e lemma). A
nonzero $d\beta$ at a state therefore rules out every function whose gradient is R/R's average update plus $c$ on a
neighbourhood of that state. Because $\beta$ has only score components, $d\beta$ has two blocks: one pairs the scores with
$V$ and holds the derivatives $\partial\delta_i/\partial V_{jk}$ themselves, and one lies within the scores.

Under the symmetric law $r_j=\pm\varepsilon$ with $\varepsilon=\tfrac18$, the variance does not depend on $p$, so $c=0$
and $\delta=\varepsilon^2\big[g-(\mathbf 1^\top g)\,p-2Gp\big]$. The largest entry of $\lvert\partial\delta/\partial V\rvert$
is \NSOneNineThreeCCellAMaxDbdV{} in cell A and \NSOneNineThreeCCellBMaxDbdV{} in cell B. It is exactly
\NSOneNineThreeCNullMaxDbdV{} in a null cell with $n=2$, $d=1$, $p=(\tfrac12,\tfrac12)$ and $v_2=-v_1$, where $\delta$
also vanishes, because $Gp=0$ and $g=(\mathbf 1^\top g)\,p$. A direct check that skips
the reduction to $\delta$ compares the mixed partial derivatives of R/R's whole average update in cell B: they differ by
up to \NSOneNineThreeCMaxGapReplay{}, while U/R's agree exactly (largest difference \NSOneNineThreeCMaxGapRole{}). Under
the symmetric law $r_j=\pm p_j/4$, the variance depends on $p$ and $c$ is not identically zero. Exact enumeration of both
blocks of $d\beta$ then gives the same pattern: nonzero in both cells, zero in the null cell. The null cell's zero belongs
to the symmetric laws: under stochastic rounding to a grid of spacing $\tfrac25$, with $v_1=1$, exact enumeration finds
$d\beta$ nonzero there. In the null cell, R/R's average errors over the forward's rounding vanish at both uses under the
symmetric laws. The theorem still holds there, because it holds the forward fixed and asks for every incoming gradient.

We use the classical criterion that a gradient field has a symmetric Jacobian, also used to analyze game dynamics
\citep{balduzzi2018mechanics}. Non-gradient updates also arise in contrastive divergence
\citep{sutskever2010convergence}. Here $\beta$ is the difference between two backward-state policies of one computation,
with the same correction $c$ for the rounding law.

Being a gradient does not decide whether a policy follows the reference. U/U's average update is exactly the gradient of
the loss without rounding, $\tfrac12p^\top Gp$, yet U/U errs at the gradient of $V$. U/R follows the reference at both
uses, yet its average update is $\nabla\bar L-c$. Under the law $r_j=\pm p_j/4$, $c$'s score part depends on $V$ while its
$V$ part is zero, so in the two cells U/R's average update is not a gradient until $c$ is added. Whether a policy is right
is decided use by use, against the reference.

%% file: sections/C-contract.tex
%
\section{The checker and the operator tests}\label{sec:app_contract}

\subsection{What the checker reads and reports}\label{sec:app_contract-checker}

For each use, the checker reads the value the policy supplies and the random draws on which its gap from the
reference and its incoming gradient depend. The policy declares whether it uses a substitute and, for a new rounding,
whether the original is kept and which rounding distributions must be supported. It also lists the possible outcomes,
on which the checker evaluates the policy and reference rules.

The verdicts rest on one split of the use's error. Write the use's backward rule as $Bh$ and the reference's as
$B^{\ast}h$, where $h$ is the incoming gradient, and let $\Delta=B-B^{\ast}$. The requirement of
Section~\ref{sec:contract-req} fixes the forward, including its rounding, and averages only over the substitute's own
draw. For the split below, we average over the forward's rounding as well, with the forward's inputs and weights
fixed. At that level the use's average error splits into two parts:
\begin{equation*}
\mathbb{E}[\Delta h] \;=\; \mathbb{E}[\Delta]\,\mathbb{E}[h] \;+\; \mathbb{E}\big[(\Delta-\mathbb{E}\Delta)\,(h-\mathbb{E}h)\big].
\end{equation*}
The first part is a mean error: the substitute is off on average. The second is a coupling: the gap moves with the
incoming gradient because both depend on one draw. A zero average at this level is weaker than the requirement itself.

The checker compares rules over the declared outcomes; it does not compute averages. Replay certifies a zero gap.
A substitute whose effect is zero in every outcome also certifies the whole error, even when a mean error and a
coupling cancel. Separate draws, independent given the forward's inputs, certify zero coupling while leaving the mean
error undecided. A shared draw receives an undecided verdict. A requirement fails when the requested procedure cannot
produce it. For example, a procedure given only a rounded value, the grid and new random bits cannot independently
redraw from the original distribution for two possible original values between the same adjacent grid points: the
only procedure that preserves both distributions returns the saved value.

\subsection{How the checker's verdicts were tested}\label{sec:app_contract-oracle}

The verdict tests use rational arithmetic and enumerate every rounding outcome, with equality comparisons and no
tolerance. Ten policy variants cover replay, independent reconstruction, restored random state, detachment from the
computational graph used for automatic differentiation, and a new rounding requested without its original. Their expected verdicts distinguish
zero total error, zero coupling with a possible mean error, an undecided shared draw, and an impossible redraw. Every
variant received its expected verdict, and exact enumeration confirmed each certificate or counterexample.

Seven families of exact checks derive their reference values from a declared forward computation. They include
recomputation under restored random state, softmax Jacobians formed from one or two independent copies, and the
impossibility of a redraw from the saved value alone. Mutation tests change the objects being checked and compare the
outcome with the expectation declared before the run. They also include an edit that preserves the average, which the check of
that average correctly accepts. Every edit landed as declared.

The contrast between shared and independent copies in Section~\ref{sec:contract-predict} reads the coupling when the reference
value is uncorrelated with the incoming gradient. In attention, the reference output $PV$ is fixed given the forward
inputs, so this condition holds. For a composed backward, each stage must be conditioned on the input it actually
receives. An exact example with two stages gives the reference on average with separate draws but doubles the mean gradient
with one shared draw, although each stage passes when tested at a fixed input. Conditioning on the received input, or
checking the whole backward against the reference, detects the error.

\subsection{How the three tests were registered}\label{sec:app_contract-protocol}

Each test registered its operator, setup, predictions and reading rules before measurement: before probe code existed
for the CPU tests, and before substituted backward calls for the attention test. The checkpoint and log-softmax tests
were repeated in a second, separately registered configuration. Our code controls the rounding or sampling; the
operators are those of PyTorch or TE, and nothing is trained. A statement \emph{held} when every part of its rule was
met and \emph{missed} otherwise. Tables~\ref{tab:app-ckpt}--\ref{tab:app-te} report every registered statement, including
LS5b, which was predicted to miss, and the two other misses, S6 and LS4.

\subsection{Checkpointing}\label{sec:app_contract-ckpt}

The first test of Section~\ref{sec:contract-tests} asks whether checkpointing's switch for the random state removes the
effect of reuse. The model computes $y=W_2\,\mathrm{Dropout}(\mathrm{gelu}(W_1x))$, with dropout rate 0.5. It rounds $y$
stochastically to a grid of step 0.05, passing gradients through the rounding unchanged, and ends in a linear layer and
a squared loss. An operation placed after the rounding records, for each row, the added term in
\eqref{eq:dterm} for a use that reads the rounded value $\tilde y$ instead of the original:
$\Delta D=\sum_j \mathrm{d}\tilde y_j\,(\tilde y_j-y_j)$, where $\mathrm{d}\tilde y$ is the incoming gradient. It passes
the gradient on unchanged. The use's mean error $m$ is the mean of $\Delta D$ over 32 rows and $n=512$ trials; its
standard error (se) is taken over trials, and $z=|m|/\mathrm{se}$. A gate, computed on a separate check of the rounding, required the
rounding's relative root-mean-square error to lie between 1\,\% and 10\,\%; the values the runs used also lie inside that
range. There are five conditions: no checkpointing (the baseline), and a checkpointed part
that either ends at the rounding or continues through the last layer and the loss, each with the random state restored
or not. The width is 256. The code runs PyTorch 2.14's non-reentrant checkpointing (\texttt{use\_reentrant} set to false),
eagerly on CPU, with its default determinism check. The second configuration has width \NSTwoTwoSixCKWidth,
$n=1024$ trials and a registered seed, in two layouts: one with the dropout, and one without it, so that only the
rounding is drawn anew.

\begin{table}[ht]
\centering
\caption{\textbf{Checkpointing: every registered statement and its outcome.} ``Removed'' and ``kept'' refer to the
use's mean error when the random state is not restored. Percentages and ratios are relative to the baseline without
checkpointing, in magnitude.}
\label{tab:app-ckpt}
\small
\begin{tabular}{@{}>{\raggedright\arraybackslash}p{0.27\linewidth}>{\raggedright\arraybackslash}p{0.27\linewidth}>{\raggedright\arraybackslash}p{0.21\linewidth}>{\raggedright\arraybackslash}p{0.16\linewidth}@{}}
\toprule
Statement & Registered rule & Measured & Outcome \\
\midrule
\multicolumn{4}{@{}l}{\emph{First configuration: width 256, $n=512$}} \\
S1. Restoring the random state makes checkpointing change nothing
  & record bit-identical to the baseline's, wherever the checkpointed part ends; masks agree exactly
  & bit-identical; masks agree & held \\
\addlinespace[3pt]
S2. Without restoration, the effect is removed when recomputation stops at the rounding
  & $z<2$ and under 10\,\%; baseline $z>5$
  & $z=\NSOneNineOneOutFZ$, \NSOneNineOneOutFPctOfBase\,\%; baseline $z=\NSOneNineOneBaseZ$ & held \\
\addlinespace[3pt]
S3. It is kept when recomputation includes the loss
  & $z>5$, the baseline's sign, over 30\,\%
  & $z=\NSOneNineOneInFZ$, same sign, \NSOneNineOneInFPctOfBase\,\% & held \\
\addlinespace[3pt]
S4. Without restoration, the dropout mask changes
  & agreement between $0.30$ and $0.85$; exactly 1 with restoration
  & \NSOneNineOneOutFMaskAgreement; all entries with restoration & held \\
\addlinespace[3pt]
S5. PyTorch's own check stays silent
  & no error or warning in any condition & none & held \\
\addlinespace[3pt]
S6. With round-to-nearest, turning restoration off changes nothing
  & record bit-identical to the baseline's; baseline $z>5$
  & not bit-identical & missed: our dropout made $y$ random \\
\addlinespace
\multicolumn{4}{@{}l}{\emph{Second configuration: width \NSTwoTwoSixCKWidth, $n=1024$}} \\
W1. As S1, with dropout
  & as S1 & bit-identical; masks agree & held \\
\addlinespace[3pt]
W2. As S2, with and without dropout
  & mean $\pm1.96\,$se inside 10\,\%; baseline $z>5$
  & \NSTwoTwoSixCKROutfPctOfBase\,\% with, \NSTwoTwoSixCKIOutfPctOfBase\,\% without; baseline $z=\NSTwoTwoSixCKRBaseZ$, \NSTwoTwoSixCKIBaseZ
  & held \\
\addlinespace[3pt]
W3. As S3, with and without dropout
  & $z>5$, the baseline's sign, ratio within $0.15$ of 1
  & ratio \NSTwoTwoSixCKRInfRatio{} with, \NSTwoTwoSixCKIInfRatio{} without & held \\
\addlinespace[3pt]
W4. As S4, with dropout
  & as S4 & \NSTwoTwoSixCKRMaskAgreeOutf; all entries with restoration & held \\
\addlinespace[3pt]
W5. As S5, both layouts & as S5 & none & held \\
\addlinespace[3pt]
W6. As S6, without dropout
  & as S6 & bit-identical; baseline $z=\NSTwoTwoSixCKRtnBaseZ$ & held \\
\bottomrule
\end{tabular}
\end{table}

Recomputing the loss couples the new draw to its own incoming gradient, giving the same expected error as the
baseline. The registered rules in Table~\ref{tab:app-ckpt} distinguish removal from retention; the second configuration
uses an equivalence test for removal and a point prediction for the retained ratio. The determinism check in PyTorch
compares shapes, data types and devices, so it does not detect changed tensor values.

S6 specified a deterministic input to the rounding, but the test included dropout, which changed the record when the
random state was not restored. The second configuration tested this control without dropout, and W6 held.

\FloatBarrier
\subsection{Log-softmax}\label{sec:app_contract-lsm}

The second test varies the loss instead of what is recomputed. PyTorch's CPU backward for log-softmax takes the
incoming gradient $h$ and the saved output $S$ and returns $h-\exp(S)\,(\mathbf{1}^{\top}h)$, where $\mathbf{1}$ is the
all-ones vector; in double precision it matches this formula exactly. Our code saves $S=\ell+r$, where $\ell$ is the
exact output and the entries $r_j$ are independent, each $\pm\varepsilon$ with equal probability. Under reuse the
backward reads this $S$; under an independent copy it reads $\ell+r'$, with $r'$ drawn anew. Write $p=\exp(\ell)$,
$g$ for the returned gradient, and $g^{\ast}$ for what the same operator returns from $\ell$ with the same $h$. The
statistic averages $(g_j-g^{\ast}_j)/p_j$ over rows and classes. The first configuration uses $\varepsilon=0.1$, 128
classes, 64 rows and 4096 trials, in double precision on CPU.

The predictions about reuse in Table~\ref{tab:app-lsm} follow from the formula. Under negative log-likelihood with a fixed target, $h=-e_y$, the negated one-hot
vector of the target, so $\mathbf{1}^{\top}h=-1$. Each entry's error divided by $p_j$ is then $e^{r_j}-1$, whose average
is $\cosh\varepsilon-1$ for either copy: reuse changes nothing on average, yet neither copy is right. Under the
constructed loss $\tfrac12\lVert S-\ell\rVert^2$, with $\ell$ held constant, the incoming gradient is $h=r$. Each entry's
error divided by $p_j$ is then $-(e^{r_j}-1)\sum_k r_k$. Under reuse its average is
$-\mathbb{E}[r_je^{r_j}]=-\varepsilon\sinh\varepsilon$; under an independent copy it is zero. Our code built this loss
so that the incoming gradient equals the rounding error. A
likelihood with a fixed target, label smoothing and distillation from a fixed teacher all give an incoming gradient that does
not depend on $S$; entropy penalties and focal losses give one that does.

\begin{table}[ht]
\centering
\caption{\textbf{Log-softmax: every registered statement and its outcome.} ``Likelihood'' is the negative
log-likelihood with a fixed target, ``constructed'' the loss whose incoming gradient equals the rounding error. Errors are per entry,
divided by the entry's probability; relative errors are against the registered constant.}
\label{tab:app-lsm}
\small
\begin{tabular}{@{}>{\raggedright\arraybackslash}p{0.28\linewidth}>{\raggedright\arraybackslash}p{0.25\linewidth}>{\raggedright\arraybackslash}p{0.22\linewidth}>{\raggedright\arraybackslash}p{0.16\linewidth}@{}}
\toprule
Statement & Registered rule & Measured & Outcome \\
\midrule
\multicolumn{4}{@{}l}{\emph{First configuration: $\varepsilon=0.1$}} \\
LS1. Under the likelihood, reuse changes nothing on average
  & $z<2$ and under 5\,\% of the constructed loss's effect
  & $z=\NSOneNineTwoLSOneZ$, \NSOneNineTwoLSOnePctOfB\,\% & held \\
\addlinespace[3pt]
LS2. Under the likelihood, the mean error is $\cosh\varepsilon-1$
  & $|\text{relative error}|<0.02$; $z>5$ & relative error \NSOneNineTwoLSTwoRelErr; $z$ above 5 & held \\
\addlinespace[3pt]
LS3. Under the constructed loss, reuse shifts the error by $-\varepsilon\sinh\varepsilon$
  & $|\text{relative error}|<0.05$; $z>5$ & relative error \NSOneNineTwoLSThreeRelErr; $z$ above 5 & held \\
\addlinespace[3pt]
LS4. Under the constructed loss, the independent copy's mean error is zero
  & $z<2$ and under 5\,\% of LS2's error
  & $z=\NSOneNineTwoLSFourZ$, \NSOneNineTwoLSFourPctOfA\,\% & missed: $z>2$ \\
\addlinespace[3pt]
LS5a. Dividing the reconstruction by $\cosh\varepsilon$ removes the likelihood's error
  & $z<2$ and under 5\,\% of it & $z=\NSOneNineTwoLSFiveACalibZ$, \NSOneNineTwoLSFiveACalibPctOfAShared\,\% & held \\
\addlinespace[3pt]
LS5b. After that division, the constructed loss's effect of reuse is zero
  & $z<2$ and under 5\,\% of the undivided effect; predicted to miss
  & ratio to the undivided effect \NSOneNineTwoLSFiveCalibRatio{} $\approx1/\cosh\varepsilon$ & missed, as predicted \\
\addlinespace[3pt]
LS6. A real bfloat16 store obeys the same formula
  & statistic equals the mean of $e^{r_j}-1$, $|\text{relative error}|<0.01$; nonzero
  & relative error \NSOneNineTwoLSSixRelErr; $z=\NSOneNineTwoLSSixZ$ & held \\
\addlinespace
\multicolumn{4}{@{}l}{\emph{Second configuration: $\varepsilon=\NSTwoTwoSixLSEps$}} \\
LS1$'$. As LS1
  & mean $\pm1.96\,$se inside $\pm$5\,\% of the constructed effect
  & mean \NSTwoTwoSixLSOneContrast; bound $\pm$\NSTwoTwoSixLSOneMargin & held \\
\addlinespace[3pt]
LS2$'$. As LS2
  & $|\text{relative error}|<0.01$; $z>5$ & \NSTwoTwoSixLSTwoRelErr; $z=\NSTwoTwoSixLSTwoZ$ & held \\
\addlinespace[3pt]
LS2$'$d. The measurement rejects the leading term $\varepsilon^2/2$
  & $|\text{relative error}|$ from $\varepsilon^2/2$ at least $0.01$ & \NSTwoTwoSixLSTwoDRelErrLead & held \\
\addlinespace[3pt]
LS3$'$. As LS3
  & $|\text{relative error}|<0.02$; $z>5$ & \NSTwoTwoSixLSThreeRelErr; $z=\NSTwoTwoSixLSThreeZ$ & held \\
\addlinespace[3pt]
LS3$'$d. The measurement rejects the leading term $-\varepsilon^2$
  & $|\text{relative error}|$ from $-\varepsilon^2$ at least $0.02$ & \NSTwoTwoSixLSThreeDRelErrLead & held \\
\addlinespace[3pt]
LS4$'$. As LS4
  & mean $\pm1.96\,$se inside $\pm$10\,\% of LS2$'$'s error
  & mean \NSTwoTwoSixLSFourBIndep; bound $\pm$\NSTwoTwoSixLSFourMargin & held \\
\addlinespace[3pt]
LS5a$'$. As LS5a
  & mean $\pm1.96\,$se inside $\pm$5\,\% of LS2$'$'s error & mean \NSTwoTwoSixLSFiveACalib & held \\
\addlinespace[3pt]
LS6$'$. As LS6 & as LS6 & relative error \NSTwoTwoSixLSSixRelErr & held \\
\bottomrule
\end{tabular}
\end{table}

For LS4, the magnitude was within its registered bound, but the z-score exceeded 2. The second configuration
registered equivalence tests for its nulls, and LS4$'$ held. LS5b was predicted to miss: dividing the reconstruction by
$\cosh\varepsilon$ corrects its mean but leaves the coupling scaled by $1/\cosh\varepsilon$.
LS6 uses a real bfloat16 store. Its calibration was revised after the first result was read so that it could fail;
the verdict was unchanged. The second configuration also distinguishes the exact constants from their leading
terms (LS2$'$d, LS3$'$d).

\FloatBarrier
\subsection{FP8 attention}\label{sec:app_contract-te}

The third test takes the contrast of Figure~\ref{fig:pairing} to TE's FP8 attention backward. It compares the gradients
this backward returns with a model in which $D$ is formed from the saved output, as in FlashAttention. The test used
TE 2.13 with cuDNN 9.20 and PyTorch 2.13 on one NVIDIA H20 GPU. TE's attention module has the attention shape of our
390M model (8 query heads, 4 key-value heads, head dimension 128, causal). It runs under the FP8 recipe of our training
runs, with FP8 attention in both passes: E4M3 in the forward and E5M2 in the backward. One pass through the module
captured the arguments of its fused attention backward. Every test call passes the same arguments to that captured
function, except three: the saved output, the incoming gradient, and the quantizers, whose scales are set to one. The
test thus calls the backward operator directly.

Our code sets the inputs so that every value that decides the result is exact. Each key's first coordinate is 1, and
each query is orthogonal to every key, so every score is zero and attention is uniform. The active row, row 63, attends
to 64 keys whose values alternate between 1 and 1.5 in the first coordinate, so its output entry there is $s=5/4$. Our
code sets that entry of the saved output to $s+\sigma a$ in the shared sample and $s+\tau a$ in the independent one,
with $a=1/4$ and signs $\sigma,\tau=\pm1$. It sets the incoming gradient $\mathrm{d}O$ to zero except at that entry,
where it is $h$. The model computes $D=\mathrm{rowsum}(\mathrm{d}O\circ O)$ from the saved output $O$, then
$\mathrm{d}S=P\circ(\mathrm{d}P-D)$ and $\mathrm{d}Q=c\,\mathrm{d}S\,K$, with the softmax scale set by our code to $c=1/8$
(the model's own is $1/\sqrt{128}$). On this case it
returns $c\,h\,(s-x)$ at the active entry of the query gradient, for a saved entry $x$. The contrast $T$, the query
gradient from the independent sample minus that from the shared one, divided by $c$, is then $h\,(\sigma-\tau)\,a$
(Table~\ref{tab:app-te}). With $h=\sigma a$, the shared sample's deviation, its mean over the four sign pairs is $a^2$;
with a fixed $h=a$, it is zero. In the same model, a $D$ that ignored the saved output would give $T=0$ in every cell.

\begin{table}[ht]
\centering
\caption{\textbf{FP8 attention: the eight predicted cells and the measurement.} $T$ is the query gradient from the
independent sample minus that from the shared one, divided by the softmax scale.}
\label{tab:app-te}
\small
\begin{tabular}{@{}lcc@{}}
\toprule
Signs $(\sigma,\tau)$ of the two samples & $h=\sigma a$ (shared sample's deviation) & $h=a$ (fixed) \\
\midrule
$(+,+)$ & $0$ & $0$ \\
$(+,-)$ & $2a^2$ & $2a^2$ \\
$(-,+)$ & $2a^2$ & $-2a^2$ \\
$(-,-)$ & $0$ & $0$ \\
\midrule
Predicted mean & $a^2=\NSTwoTwoSevenGpurowPredMeanValue$ & $0$ \\
Measured mean & \NSTwoTwoSevenGpurowMeanValue & \NSTwoTwoSevenGpurowMeanFixed \\
\midrule
Registered rule & \multicolumn{2}{c}{every cell within $a^2/10=\NSTwoTwoSevenGpurowEps$ of its prediction} \\
Largest deviation of a cell & \multicolumn{2}{c}{\NSTwoTwoSevenGpurowMaxAbsDev{} over all eight cells: held} \\
\bottomrule
\end{tabular}
\end{table}

Before computing the contrast, checks confirmed the software and Hopper GPU, the FP8 fused backend and inputs, and
unit scales on all tensors and quantizers. The decoded saved output and incoming gradient had to match the intended
values, and the value gradient had to show uniform probabilities and be byte-identical within each pair. Repeated
calls also had to return byte-identical gradients. All checks passed. The same test rejected an emulated backward
whose $D$ ignores the saved output. Exact representation of the active forward and backward values made zero
deviation possible. The test observes returned gradients; $D$ and the path through the saved output belong to the
model that predicts them.

\FloatBarrier

%% file: sections/D-setup.tex
%
\section{Setup}\label{sec:app_setup}

This appendix specifies the training runs and probes in Sections~\ref{sec:outcome} and~\ref{sec:uses}.
Appendix~\ref{sec:app_contract} gives the operator tests, and Appendix~\ref{sec:app_evidence} the additional runs.

\subsection{Model, data and software}\label{sec:app_setup_model}

We use the FOG setting of Llama~3 \citep{grattafiori2024llama,hernndezcano2025fully}, at 390M parameters except for
one 1.5B comparison (Table~\ref{tab:app-recipe}). The pre-norm decoder uses root-mean-square normalization \citep{zhang2019root},
rotary position embeddings \citep{su2021roformer}, grouped-query attention \citep{ainslie2023gqa} and a SwiGLU MLP
\citep{shazeer2020glu}, without biases. We initialize weights from $\mathcal{N}(0,0.02^2)$, using
$\mathcal{N}(0,0.02^2/2L)$ for the two output projections of each of the $L$ blocks, and set normalization gains to one.
Head dimension is 128 and softmax scale is $1/\sqrt{128}$ at 390M.

\begin{table}[t]
\centering
\caption{\textbf{Training recipes.} ``Decay'' is the length of the final decay; ``none'' holds the peak rate to the end.
The forks of Sections~\ref{sec:outcome-forks} and~\ref{sec:outcome-output} use the first column's model and batch
on their own schedules (Appendices~\ref{sec:app_setup_forks} and~\ref{sec:app_evidence}).}
\label{tab:app-recipe}
\small
\setlength{\tabcolsep}{4pt}
\begin{tabular}{@{}lccc@{}}
\toprule
 & 390M, 20k schedule & 390M, store runs & 1.5B \\
\midrule
Used in & Sections~\ref{sec:outcome-backward}--\ref{sec:outcome-output} & Section~\ref{sec:uses} & Section~\ref{sec:outcome-backward} \\
Layers; width; MLP width & 16; 1024; 4096 & 16; 1024; 4096 & 16; 2048; 8192 \\
Query heads; key-value heads & 8; 4 & 8; 4 & 16; 8 \\
Input and output embeddings & tied & tied & untied \\
Sequences $\times$ tokens per step & $128\times4096$ & $128\times4096$ & $256\times4096$ \\
GPUs $\times$ micro-batch $\times$ accumulation & $8\times4\times4$ & $8\times4\times4$ & $8\times4\times8$ \\
Peak learning rate & $10^{-3}$ & $10^{-3}$ & $2.5\times10^{-4}$ \\
Warmup; decay; total steps & 5000; 4000; 20000 & 2000; 1600; 8000 & 2500; none; 5000 \\
Rounding below BF16 & FP8 or none, per run & 6- or 5-bit store & FP8 or none, per run \\
Seeds & 0, 1, 2 & 0, 1 & 0 \\
\bottomrule
\end{tabular}
\end{table}

We use the 10B-token FineWeb-Edu sample \citep{penedo2024fineweb} with the FOG Mistral-NeMo tokenizer (131,072 tokens),
one beginning-of-sequence token per document and no end-of-document token. We hold out the first 5000 documents of the
last shard. Training reads non-overlapping 4096-token windows in an order fixed by the seed; the seed also sets initialization.
Validation (held-out) loss averages the same 128 held-out windows every 250 steps and at the last step.

AdamW \citep{loshchilov2017decoupled} uses $\beta_1=0.9$, $\beta_2=0.95$, $\epsilon=10^{-8}$, weight decay 0.1 on weight
matrices, gradient clipping at norm 1.0, BF16 autocast and FP32 master weights. The learning rate warms up linearly,
holds its peak, then decays toward $10^{-8}$ as $1-\sqrt{x}$, where $x$ is the elapsed fraction of the decay
\citep{hu2024minicpm,hagele2024scaling}. Our 20k schedule retains the 5000-step warmup of the FOG 100k schedule and
shortens its total length and decay by a factor of five.

FP8 runs use TE delayed scaling: HYBRID format (E4M3 forward, E5M2 gradients \citep{micikevicius2022formats}), margin 0,
and the maximum of 1024 past absolute maxima. All block linear layers run in FP8. FP8 attention uses the fused forward
(\texttt{fp8\_dpa} on, \texttt{fp8\_mha} off) and backward (\texttt{NVTE\_FP8\_DPA\_BWD=1}), except where the backward
is replaced. Embeddings and the output head stay in BF16 and FP32. BF16 runs disable FP8 in the same model. All training
is data-parallel on 8 GPUs of one node; the recorded GPU model is NVIDIA H20.

Table~\ref{tab:app-images} records the software builds. Comparisons on newer software are reported separately, with their
references identified in Appendix~\ref{sec:app_evidence_scratch}. The CPU operator tests used PyTorch 2.14
\citep{paszke2019pytorch}.

\begin{table}[t]
\centering
\caption{\textbf{Software builds.} PyTorch and TE build strings, with cuDNN versions where recorded.}
\label{tab:app-images}
\footnotesize
\begin{tabular}{@{}l l l l >{\raggedright\arraybackslash}p{0.35\linewidth}@{}}
\toprule
Build & PyTorch & TE & cuDNN & Runs \\
\midrule
Earlier & 2.11.0+cu128 & 2.10.0+769ed778 & 9.16 & the forks of Figure~\ref{fig:pairing}a; every FP8 run of
  Figure~\ref{fig:backward-only} and its seed-0 and seed-1 BF16 runs; the runs of Section~\ref{sec:outcome-output}; the
  1.5B runs \\
Newer (cu129) & 2.13.0+cu129 & 2.13.0+28777046 & not named & the seed-2 BF16 run of Figure~\ref{fig:backward-only}; a
  seed-0 BF16 repeat; the 6-bit store runs \\
Newer (cu130) & 2.13.0+cu130 & 2.13.0+28777046 & 9.20 & Figure~\ref{fig:pairing}b; the 20k runs on newer software
  (Section~\ref{sec:outcome-backward}) and their own BF16 runs; the 5-bit store runs; the continuations of
  Section~\ref{sec:uses-loss}; the TE test of Section~\ref{sec:contract}, on one H20 \\
\bottomrule
\end{tabular}
\end{table}

\subsection{The forks of Section~\ref{sec:outcome-forks}}\label{sec:app_setup_forks}

Each fork resumes weights, optimizer and random-number states at step 999 of a 390M run with TE FP8 attention, on seed
0 or 1. The parents use the first recipe in Table~\ref{tab:app-recipe}, with a 5000-step warmup and no decay. Forks run
updates 1000 to 3999, within the warmup.

The next layer consumes a stochastic rounding of $O$, recomputed as $PV$ in FP32 from the dequantized FP8 $Q$, $K$ and $V$
with $P$ unrounded. Each entry is rounded between neighbouring E4M3 values at the TE output scale, using their BF16
values to set probabilities so that the consumed rounding averages to $O$. Unbracketed entries at the top of the range
round to nearest and are counted; any other unbracketed entry invalidates the run. A counter-based generator is keyed
on rounding seed, stream, rank, step, micro-batch and layer. The new rounding uses a separate stream of the same law.
Evaluation reads the original TE output.

The forks use a dense FP32 emulation of the TE 2.10 FP8 attention backward, whose inspected code computes $D$ from the
saved output. The emulation casts $dO$ to E5M2, recomputes $P$, and forms $dV=P_q^\top dO$, where $P_q$ is rounded at the
forward scale. It computes $dP=dO\,V^\top$, takes $D$ from the assigned output copy or from
$\mathrm{rowsum}(P\circ dP)$, and forms $dS=P\circ(dP-D)$. Scaled $dS$ and the gradients of $Q$, $K$ and $V$ are cast to
E5M2 before the gradients return in BF16. Three delayed scales, initialized from TE backward history, serve $dO$, $dS$
and the three gradients. Among the cast placements compared, this one is closest to the TE gradients: over 16 layers
on one held-out window at seed-0 steps 999 and 1999, relative distances were
\NSTwoTwoNineEmulDqMin--\NSTwoTwoNineEmulDqMax\% for $Q$ and
\NSTwoTwoNineEmulDkMin--\NSTwoTwoNineEmulDkMax\% for $K$.

Runs that reuse the rounding and those that read a new one differ only in the tensor read by $D$, at fixed inputs, incoming gradient and scales.
The three pairs combine parent
seed 0 with rounding seeds 1 and 2, and parent seed 1 with rounding seed 1. The two pairs with rounding seed 1 share a stream;
the pair with rounding seed 2 uses the control with rounding seed 1. Figure~\ref{fig:pairing}b repeats the first pair on newer
software.

The evaluation rule was fixed before launch. Any non-finite loss or excursion more than 1.5 nats above
the running minimum fails a fork. A useful run must also meet three endpoint conditions: attention logits below
$10^3$ at every observed probe in steps 3400 to 3999, with at least 8 of 12 probes complete and finite in all layers;
mean training loss over steps 3800 to 3999 at most 0.05 nats above the control; and the same bound on validation loss at
step 3999. Logits are probed every 50 steps on the first 1024 tokens of one sequence. All loss bounds are one-sided.

\subsection{TE attention trained from scratch}\label{sec:app_setup_te}

The 390M runs of Figure~\ref{fig:backward-only} and Section~\ref{sec:outcome-output} train from scratch on the 20k
schedule, with seeds 0, 1 and 2. The BF16 backward captures dequantized FP8 $Q$, $K$ and $V$ through a separate quantizer
using the current TE scale. It recomputes the output and softmax normalizer with FlashAttention-2 in BF16, then runs
its BF16 backward \citep{dao2023flashattention2}. The 1.5B comparison uses the same construction.

Replacing the saved output keeps the TE FP8 backward. A Python wrapper substitutes stochastic E4M3 rounding of a
BF16 FlashAttention-2 recomputation from the dequantized FP8 inputs, at the scale of the saved output. All other arguments
pass unchanged. The wrapper accepts only FP8 calls with E4M3 saved output
and draws from a generator private to each run and rank.

The 20k evaluation rule reads loss every 100 steps, with the same excursion threshold. An excursion fails
only if it remains open at the end; it closes at a later reading below 0.75 nats above its running minimum. Any logged
non-finite loss fails the run. Useful runs require logits below $10^3$ on all observed probes in steps 19000 to 19999,
with at least 18 of 20 probes complete and finite in all layers, and gaps at most 0.02 nats to the BF16 reference with the same seed in both
mean training loss over steps 19800 to 19999 and validation loss at step 19999.

The 1.5B runs continue their own 3000-step jobs from step 2999 to 4999, restoring optimizer and random-number states.
Data order depends on the step. At restart, an extra evaluation enters FP8 scaling history, rank 0 random-number state is
restored on every rank, and runtime caches restart. Continuity requires the first ten losses to be finite and their
mean within 0.05 nats of the preceding ten. The rule uses the combined 100-step grid, omits the logit condition, and
allows gaps of 0.05 nats to BF16 in validation and mean training loss over steps 4800 to 4999.

Reference runs must complete without failure and, where the rule requires it, keep logits below $10^3$. A stop at the time limit
before completion leaves success undecided but preserves a failure already observed.

\subsection{The normalization store}\label{sec:app_setup_store}

Section~\ref{sec:uses} stores the input of the first MLP projection after normalization in every layer. It computes
$z$, $u=\gamma\circ z$, $u_q$ and $y=u_qW$ in FP32. For each token, the $b$-bit grid has step
$s=\max_j|u_j|/(2^{b-1}-1)$, and $u_q=s\lfloor u/s+\xi\rfloor$ with independent uniform $\xi$ on $[0,1)$ per entry.
Thus $u_q$ averages to $u$ without clipping. The new rounding $u_f=u+r_f$ uses a separate generator stream. All policies
at one seed share store random numbers.

The backward passes $du=dy\,W^\top$ through the rounding and returns $dW=u_W^\top dy$ and
$d\gamma=\sum_{\mathrm{tokens}} z_g\circ du$, with $z_g$ and $u_W$ from Table~\ref{tab:app-policies}. The normalization
input gradient uses unrounded values.

\begin{table}[t]
\centering
\caption{\textbf{Store policies.} The value each use reads, and the store widths at which each policy was trained. The
code names the gain use first. BF16 is the model trained without the store.}
\label{tab:app-policies}
\small
\begin{tabular}{@{}lcccc@{}}
\toprule
Policy & Gain use reads ($z_g$) & Weight use reads ($u_W$) & 6 bits & 5 bits \\
\midrule
U/R (reference) & $z$ & $u_q$ & \checkmark & \checkmark \\
U/U & $z$ & $u$ & \checkmark & \checkmark \\
U/N & $z$ & $u_f$ & & \checkmark \\
R/R & $u_q/\gamma$ & $u_q$ & \checkmark & \checkmark \\
R/N & $u_q/\gamma$ & $u_f$ & \checkmark & \\
N/R & $u_f/\gamma$ & $u_q$ & \checkmark & \\
N/N & $u_f/\gamma$ & $u_f$ & \checkmark & \\
U/R+$\varepsilon$ & $z$ & $u_q+(u_f-u)$ & \checkmark & \checkmark \\
BF16 & -- & -- & \checkmark & \\
\bottomrule
\end{tabular}
\end{table}

The probe of Section~\ref{sec:uses-norm} inserts a 6-bit store in all layers of the seed-0 and seed-2 BF16 models at
step 19999. All policies share one forward and backward; automatic differentiation checks each use. For each layer,
768 batches of one window each estimate the mean U/U error at the weight use, $-r^\top dy$. The statistic
$R=\|\bar m\|^2/\sum_j \mathrm{se}_j^2$, where $\bar m$ is the mean error and $\mathrm{se}_j$ the standard error of entry $j$.
It is near one at zero mean and increases with a resolved mean error.

The preregistered rule requires U/U to exceed, in at least 13 of 16 layers, the largest layerwise $R$ of a zero-mean
control. That control saves the store but multiplies unrounded $u$ in the forward, making the incoming gradient
independent of rounding. A second comparison, added after reading the results, uses the zero-mean error $-r_f^\top dy$ from the same forward:
on both models, the smallest U/U statistic exceeds the largest control statistic. U/U has exactly zero error at the gain use.

Store runs train in BF16 for 8000 steps on seeds 0 and 1, with the store as the only rounding below BF16. The 6-bit
experiment also trains without the store. Primary endpoints are validation losses at step 7999 on the same 128 windows,
with the store disabled. For each width, we fit $Y_{s,a}=\mu+\alpha_a+\beta_s+e_{s,a}$ with independent normal errors of one
variance. A policy contrast averages $Y_{s,a}-Y_{s,b}$ over the $S=2$ seeds. Its standard error is
$\hat\sigma\sqrt{2/S}$, and its 95\% $t$ interval has $(A-1)(S-1)$ degrees of freedom for $A$ policies:
\NSTwoOneOneDf{} at 6 bits and \NSTwoOneFiveDoseFiveDf{} at 5 bits. Intervals within a width share the variance estimate.
Equivalence requires the whole interval inside $\pm$\NSTwoOneOneDeltaNats{} nats; a difference requires exclusion of
zero. Both may hold. At 5 bits, an interval wholly beyond the margin is also labelled material.

The margin was fixed before the 6-bit endpoints were read and retained before the 5-bit runs. It was chosen from a
fifth of an earlier measurement of the 6-bit store loss cost at one model state and a small multiple of the gap between repeated runs.
The named 6-bit comparisons are U/U, U/R+$\varepsilon$ and R/R against U/R; U/R+$\varepsilon$ against U/U; U/U against
N/N; U/R against N/R; and R/N against R/R. The 5-bit comparisons are U/U, U/R+$\varepsilon$, U/N and R/R against U/R,
and U/R+$\varepsilon$ and U/N against U/U. Each comparison was named before its endpoints were read.

As a sensitivity analysis, we also report 95\% paired $t$ intervals with standard error $|D_0-D_1|/2$ and one degree of
freedom, where $D_s=Y_{s,a}-Y_{s,b}$. The final result data include pooled and paired intervals for every named comparison,
evaluated with and without the store; the pooled model above remains the primary analysis. For 6-bit U/U$\,-\,$U/R,
the paired interval lies inside the fixed margin under either evaluation mode.

The continuations of Section~\ref{sec:uses-loss} resume the two runs without the store at step 5999 for 750 updates under U/R,
U/U, R/R, or a control that computes but does not consume the 6-bit store. They share the store random numbers, retain
the schedule (decay from step 6400), and use a later revision of the training code. We probe the starting state and steps
6249, 6499 and 6749 with the same store.

At the last layer, two forwards share earlier layers and draw stores with errors $r_a$ and $r_b$ and incoming gradients
$dy_a$ and $dy_b$. Conditional on the earlier layers,
$X=-\frac12(r_a-r_b)^\top(dy_a-dy_b)$ averages to the error at the weight use $-r^\top dy$.
Products of $X$ across 472 distinct held-out windows estimate its squared mean norm. For independent identically
distributed windows this estimator is unbiased; on the fixed windows its downward bias is approximately the
variance of the average error across windows divided by the number of windows. State differences use 95\% paired
jackknife intervals over the same 8 blocks of windows (7 degrees of freedom).

The rules were fixed before reading the probes and apply to both seeds. At step 6749, attenuation requires the U/R
and U/U squared norms each to be below the control with intervals excluding zero. The primary U/R rule also tests
whether its squared norm falls below half the control after checking for a resolved increase; the comparison of U/U with half the control
is descriptive. Persistence requires $-\langle X,W\rangle/\|W\|$ to have a positive mean with $t\ge2$ under U/R and U/U.
Gain drift requires the mean layerwise log ratio of R/R to U/R root-mean-square gains to be negative at step 6249 and at most
$\log 0.95$ at step 6749, with R/R below U/R in at least 12 layers at the latter step. A fall by the first probe also
requires R/R below its own starting gain in at least 12 layers at both steps. At step 6249 it is below U/R in
\NSTwoSeventeenOnsetSZeroBelowRefSixTwoFourNine{} layers on seed 0 and
\NSTwoSeventeenOnsetSOneBelowRefSixTwoFourNine{} on seed 1. A probe state is void if either zero-mean control using a new rounding reaches $|t|>3.5$. The rule requires complete
continuations and a control that does not consume the store; a missing or void state leaves it undecided.

\subsection{Compute}\label{sec:app_setup_compute}

The \NSTwoThreeFourGpuhNRuns{} training runs underlying these comparisons took \NSTwoThreeFourGpuhTotal{} GPU-hours, counting each
run once and including the prefixes of \NSTwoThreeFourGpuhNParents{} additional parents up to their latest fork.
GPU-hours are logged time between training steps times \NSTwoThreeFourGpuhWorld{} GPUs, including intervening evaluation
and checkpointing. A 390M 20k run took a median of \NSTwoThreeFourGpuhCfgMInitTwentyKMedian{} GPU-hours
(\NSTwoThreeFourGpuhCfgMInitTwentyKSmallest{}--\NSTwoThreeFourGpuhCfgMInitTwentyKLargest{}), depending on the attention
path. This accounting excludes start-up work and work after each segment, test launches, voided or unused runs, and
probes and operator tests that do not train the model.

%% file: sections/E-evidence.tex
%
\section{Further evidence}\label{sec:app_evidence}

We report the additional comparisons cited in Section~\ref{sec:outcome} and a normalization probe of the same
rounding dependence. Appendix~\ref{sec:app_setup} specifies the training recipes and evaluation rules.

\subsection{Runs from scratch and on newer software}\label{sec:app_evidence_scratch}

Table~\ref{tab:app-scratch} gives the results for each seed behind Sections~\ref{sec:outcome-backward}
and~\ref{sec:outcome-output}, including the fork on newer software in Figure~\ref{fig:pairing}b. The runs from scratch
use the rule of Appendix~\ref{sec:app_setup_te}; the 1.5B rule permits a 0.05-nat endpoint gap and has no logit
condition. The fork on newer software uses the rule of Appendix~\ref{sec:app_setup_forks}.

\begin{table}[ht]
\centering
\caption{\textbf{Further runs of Section~\ref{sec:outcome}.} For runs that train usefully, the gaps in nats to the
block's reference: late training loss (the mean over the last 200 steps) and final validation loss. For failing runs,
the step, counted from the start of training, at which the excursion that fails the run opens.}
\label{tab:app-scratch}
\small
\begin{tabular}{@{}llccc@{}}
\toprule
Run & Seed & Outcome & Late training & Validation \\
\midrule
\multicolumn{5}{@{}l}{\emph{390M, 20k schedule, earlier software; reference: BF16 run of the same seed (seed 2's: cu129 build)}} \\
TE's FP8 attention & 0, 1, 2 & fails at \NSOneSevenFiveSZeroNativeOnset, \NSOneSevenFiveSOneNativeOnset, \NSOneSevenFiveSTwoNativeOnset & & \\
BF16 backward & 0 & useful & \NSOneSevenFiveSZeroConsDEtwoDisp & \NSOneSevenFiveSZeroConsDValDisp \\
 & 1 & useful & \NSOneSevenFiveSOneConsDEtwoDisp & \NSOneSevenFiveSOneConsDValDisp \\
 & 2 & useful & \NSOneSevenFiveSTwoConsDEtwoDisp & \NSOneSevenFiveSTwoConsDValDisp \\
Replacement of saved output & 0 & useful & \NSOneEightFourPPoSZeroDEtwo & \NSOneEightFourPPoSZeroDValDisp \\
 & 1 & useful & \NSOneEightFourPPoSOneDEtwo & \NSOneEightFourPPoSOneDValDisp \\
 & 2 & useful & \NSOneEightFourPPoSTwoDEtwo & \NSOneEightFourPPoSTwoDValDisp \\
\midrule
\multicolumn{5}{@{}l}{\emph{1.5B, continued from step 2999 to 4999; reference: BF16 run}} \\
TE's FP8 attention & 0 & fails at \NSOneSevenZeroEFpEightDpaExcursionStep & & \\
BF16 backward & 0 & useful & \NSOneSevenZeroEConsETwo & \NSOneSevenZeroEConsEThree \\
\midrule
\multicolumn{5}{@{}l}{\emph{390M, 20k schedule, cu130 build; reference: BF16 run of the same seed and build}} \\
TE's FP8 attention & 0, 2 & fails at \NSTwoTwoFourBOneSZeroEOneOnset, \NSTwoTwoFourBOneSTwoEOneOnset & & \\
BF16 backward & 0 & useful & \NSTwoTwoFourBTwoSZeroConsDeTwo & \NSTwoTwoFourBTwoSZeroConsDval \\
 & 2 & useful & \NSTwoTwoFourBTwoSTwoConsDeTwo & \NSTwoTwoFourBTwoSTwoConsDval \\
\midrule
\multicolumn{5}{@{}l}{\emph{390M fork at step 999, cu130 build (Figure~\ref{fig:pairing}b); reference: the control, whose $D$ reads no copy of $O$}} \\
Reuse & 0 & fails at \NSTwoTwoFourCIZeroSrOneSOneOpen & & \\
New rounding & 0 & useful & \NSTwoTwoFourCIZeroSrZeroGapETwo & \NSTwoTwoFourCIZeroSrZeroGapEThree \\
\bottomrule
\end{tabular}
\end{table}

For the 1.5B FP8 run, the failing excursion opens on the 100-step grid at step
\NSOneSevenZeroEFpEightDpaExcursionStep{} and remains open at the last grid reading. Its loss is
\NSOneSevenZeroEFpEightDpaLossAtFourNineZeroZero{} there, against a running minimum of
\NSOneSevenZeroEFpEightDpaRunningMinBeforeFourNineZeroZero{}, and
\NSOneSevenZeroEFpEightDpaLossAtFourNineNineNine{} at the final step 4999, before the next grid reading.

Seed 2 in Figure~\ref{fig:backward-only} uses a cu129 BF16 reference; the other runs in that figure use the earlier
build. Repeating seed 0 BF16 on cu129 changes late training loss by $\NSTwoTwoNineImgShiftETwo$ nats and validation loss
by $\NSTwoTwoNineImgShiftVal$. Shifting the seed-2 gaps by these amounts in the widening direction keeps both within
0.02 nats. The cu130 comparisons use BF16 references with the same seed and build.

\subsection{The fork at step 1999}\label{sec:app_evidence_fork}

The further fork of Section~\ref{sec:outcome-output} resumes the seed-0 parent at step 1999 for 1000 updates. The forward
consumes the original TE output rounded to nearest. Table~\ref{tab:app-fork} separates two constructions, each with its
own reference and rule, and a control with a reflected output.

\begin{table}[ht]
\centering
\caption{\textbf{Runs forked at step 1999, grouped by construction.} Labels by each block's rule; a failing run's
training loss had an excursion. Late max logit: the largest attention logit over all layers in steps 2400 to 2999. Gaps
in nats to the block's reference: mean training loss over steps 2800 to 2999, and validation loss at step 2999.}
\label{tab:app-fork}
\small
\begin{tabular}{@{}lcccc@{}}
\toprule
What the backward reads & Label & Late max logit & Late training & Validation \\
\midrule
\multicolumn{5}{@{}l}{\emph{(a) TE's FP8 backward, through the wrapper; reference: the emulated backward with $D$ from $P$}} \\
TE's output, passed through & fails & \NSOneEightFourVPassLM & \NSOneEightFourVPassETwo & \NSOneEightFourVPassEThree \\
New rounding of $O'$, rounding seed 1 & useful & \NSOneEightFourVSrLM & \NSOneEightFourVSrETwo & $\NSOneEightFourVSrEThree$ \\
New rounding of $O'$, rounding seed 2 & useful & \NSOneEightFourVSrTwoLM & \NSOneEightFourVSrTwoETwo & $\NSOneEightFourVSrTwoEThree$ \\
Nearest rounding of $O'$ & fails & \NSOneEightFourVRtnLM & \NSOneEightFourVRtnETwo & \NSOneEightFourVRtnEThree \\
Reference & & \NSOneEightFourVBrefLM & & \\
\midrule
\multicolumn{5}{@{}l}{\emph{(b) Emulated backward, what $D$ reads; reference: $D$ from $P$}} \\
TE's output & \MakeLowercase{\NSTwoTwoNineRowsPbLabelNat} & \NSTwoTwoNineRowsPbLmNat & \NSTwoTwoNineRowsPbETwoNat & \NSTwoTwoNineRowsPbEThreeNat \\
New rounding of $PV$, rounding seed 1 & useful & \NSOneEightOneIIEightLM & \NSOneEightOneIIEightETwo & \NSOneEightOneIIEightEThree \\
New rounding of $PV$, rounding seed 2 & useful & \NSOneEightOneIIEightBLM & \NSOneEightOneIIEightBETwo & \NSOneEightOneIIEightBEThree \\
$D$ from $P$ on saturated rows only & \MakeLowercase{\NSTwoTwoNineRowsPbLabelSat} & \NSTwoTwoNineRowsPbLmSat & \NSTwoTwoNineRowsPbETwoSat & \NSTwoTwoNineRowsPbEThreeSat \\
$D$ from $P$ on the other rows only & \MakeLowercase{\NSTwoTwoNineRowsPbLabelUnsat} & \NSTwoTwoNineRowsPbLmUnsat & \NSTwoTwoNineRowsPbETwoUnsat & \NSTwoTwoNineRowsPbEThreeUnsat \\
$PV$ stored in BF16 & \MakeLowercase{\NSTwoTwoNineRowsPbLabelSOneSix} & \NSTwoTwoNineRowsPbLmSOneSix & \NSTwoTwoNineRowsPbETwoSOneSix & $\NSTwoTwoNineRowsPbEThreeSOneSix$ \\
Reference & & \NSTwoTwoNineRowsPbLmB & & \\
\midrule
\multicolumn{5}{@{}l}{\emph{(c) Emulated backward, forward unreflected or reflected; reference: $D$ from $P$, unreflected}} \\
TE's output, unreflected & \MakeLowercase{\NSOneSevenSixCMNatLabel} & \NSOneSevenSixCMNatLM & \NSOneSevenSixCMNatGapETwo & \NSOneSevenSixCMNatGapEThree \\
The reflected output & \MakeLowercase{\NSOneSevenSixCMMirLabel} & \NSOneSevenSixCMMirLM & \NSOneSevenSixCMMirGapETwo & \NSOneSevenSixCMMirGapEThree \\
$D$ from $P$, reflected & \MakeLowercase{\NSOneSevenSixCMMibLabel} & \NSOneSevenSixCMMibLM & \NSOneSevenSixCMMibGapETwo & \NSOneSevenSixCMMibGapEThree \\
Reference & & \NSOneSevenSixCMCnsLM & & \\
\bottomrule
\end{tabular}
\end{table}

In construction (a), the TE FP8 backward receives either the original output through the wrapper, a stochastic E4M3
rounding of a BF16 recomputation $O'$ (rounding seeds 1 and 2), or the nearest rounding of $O'$, all at the scale of the saved
output. The reference uses an emulated backward with $D$ from $P$ and casts at higher precision. Its rule uses the fork
windows in Table~\ref{tab:app-fork}, without the logit condition: any loss excursion fails; useful runs have both loss
gaps at most 0.05 nats. Both stochastic copies train usefully; the original copy and the copy rounded to nearest fail.

Construction (b) keeps the TE FP8 forward and uses the emulated backward of Appendix~\ref{sec:app_setup_forks}.
Only the value read by $D$ changes: the TE output or a stochastic E4M3 rounding of FP32 $PV$, with rounding seeds 1
and 2. The reference computes $D$ from $P$. This rule also requires late maximum logits below $10^3$; a run exceeding
that threshold without an excursion is labelled ignited. Both stochastic copies and $PV$ stored in BF16 train usefully.

The error measurements use the first stochastic copy at the fork, in layer 10. In construction (a), its root-mean-square
error against $O'$ at the first backward call is \NSOneEightFourVSrRmsRatioLTen{} times the original TE error, with
correlation \NSOneEightFourVSrCorrLTen{}; nearest rounding has ratio \NSOneEightFourVRtnRmsRatioLTen{} and correlation
\NSOneEightFourVRtnCorrLTen{}. In (b), on rows with maximum attention probability at least 0.9, the stochastic error
against the FP32 output has ratio \NSOneEightOneIEightRmsRatio{} and correlation \NSOneEightOneIEightCorrIEightOEight{}.
The two constructions are measured separately.

For control (c), write the TE output as $O+r$, with $O$ recomputed in FP32. The forward consumes either $O+r$ or the
reflected $O-r$, cast to BF16; $D$ reads the consumed value or is computed from $P$. The reference uses $D$ from $P$
with the unreflected forward, and the rule is that of (b). The two unreflected runs repeat the corresponding (b)
configurations with identical losses and logits. Reading the reflected output
raises late logits above $10^3$ without a loss excursion over these 1000 steps; computing $D$ from $P$ trains usefully
under either forward.

\subsection{Repairing selected attention rows}\label{sec:app_evidence_rows}

The two runs that repair selected rows in Table~\ref{tab:app-fork}b recompute the selection at each backward call. A row is saturated
when its largest attention probability is at least 0.9. Computing $D$ from $P$ only on saturated
rows, while reading the TE output elsewhere, trains usefully; replacing it only on the other rows fails. The first
repair also meets the preregistered condition of no excursion and a late maximum logit at most a tenth of that in the
run that reads the TE output on every row. At the fork, saturated rows occupy at most \NSTwoTwoNineRowsPvFourMaxSatShare\% of any layer in
a separate probe. This comparison identifies a sufficient subset for repair in this fork.

\subsection{The same dependence in a normalization backward}\label{sec:app_evidence_norm}

A normalization backward can reconstruct its normalized input from the saved output, as offered by the
memory-efficient LayerNorm and RMSNorm option in NVIDIA Apex \citep{apex2023memefficient}. If the forward consumes
$u+r$, where $u=\gamma\circ z$, and passes $du$ through the rounding, this reconstruction adds
$\sum_{\mathrm{tokens}}(r/\gamma)\circ du$ to the gain gradient. Its inner product with $\gamma$ is $\langle du, r\rangle$,
whose mean can depend on pairing the rounding with the gradient.

We probe the final normalization of the seed-0 BF16 model at step 19999, using 32 held-out 4096-token windows, with
no training and an exact gradient with respect to the normalization input. The output is stochastically rounded to BF16 or to E4M3 at
one scale for the whole tensor, then dequantized and cast to BF16. For independent draws $r$ and $r'$, let $g(v)$ be the incoming
gradient when the forward consumes $v$. The paired statistic
$T=\tfrac12\langle r-r',\,g(u+r)-g(u+r')\rangle$ averages to the difference between errors from reading the consumed copy and an independent
copy. The control for the independent copy is $\langle g(u+r), r'\rangle$.

At the BF16 store the control mean is \NSOneEightSixCtrlZ{} standard errors from zero, within the registered
band of three standard errors. Moving to E4M3 multiplies median $T$ by \NSOneEightSixScaleMeasuredRatio{}; the squared ratio
of mean windowwise root-mean-square rounding errors predicts \NSOneEightSixScalePredictedRmsTwo{}, agreeing within a
factor of \NSOneEightSixScaleAgreement{}. BF16 dependence is detected by the paired
statistic; its unpaired statistic for the consumed copy remains within three standard errors of zero. The registered
test for a positive sign in each window is unresolved: \NSOneEightSixSignResolvedPositive{} of 32 BF16 windows and
\NSOneEightSixSignEResolvedPositive{} E4M3 windows exceed its tolerance. These probes use stochastic rounding;
round-to-nearest gives identical forwards and $T$ is exactly zero.